\documentclass[lettersize,journal]{IEEEtran}
\usepackage{amsmath,amsfonts}
\usepackage{algorithmic}
\usepackage{algorithm}
\usepackage{array}
\usepackage{textcomp}
\usepackage{stfloats}
\usepackage{url}
\usepackage{verbatim}
\usepackage{graphicx}
\usepackage{cite}
\usepackage{hyperref}
\hypersetup{colorlinks,
	linkcolor=blue,%
	citecolor=blue}
\usepackage[capitalize,noabbrev]{cleveref}
\usepackage{subcaption}
\usepackage{wrapfig}
\usepackage[flushleft]{threeparttable}
\usepackage{booktabs}       
\usepackage{multirow}
\usepackage[table,xcdraw]{xcolor}

\begin{document}

\title{Instance-Conditioned Adaptation for Large-scale \\ Generalization of Neural Routing Solver}


\author{
        Changliang Zhou, 
        Xi Lin, 
        Zhenkun Wang,~\IEEEmembership{Senior Member,~IEEE,}
        Xialiang Tong, 
        Mingxuan Yuan,~\IEEEmembership{Senior Member,~IEEE,} 
        Qingfu Zhang,~\IEEEmembership{Fellow,~IEEE}
\thanks{This article has been accepted for publication in IEEE Transactions on Intelligent Transportation Systems. The final published version is accessible at \url{https://doi.org/10.1109/TITS.2026.3674538}. This is the authors' self-archived version for wide dissemination.}
\thanks{C. Zhou and X. Lin are co-first authors of the article. This work was supported by the National Natural Science Foundation of China (Grant No. 62476118), the Guangdong Provincial Key Laboratory of Fully Actuated System Control Theory and Technology (Grant No. 2024B1212010002), the Natural Science Foundation of Guangdong Province (Grant No. 2024A1515011759), and the Center for Computational Science and Engineering at Southern University of Science and Technology. (\textit{Corresponding author: Zhenkun Wang}.)}
\thanks{C. Zhou and Z. Wang are with Guangdong Provincial Key Laboratory of Fully Actuated System Control Theory and Technology, School of Automation and Intelligent Manufacturing, Southern University of Science and Technology, Shenzhen 518055, China (e-mail: zhoucl2022@mail.sustech.edu.cn, wangzk3@sustech.edu.cn).}
\thanks{X. Lin is with the School of Mathematics and Statistics, Xi'an Jiaotong University, Xi'an 710049, China. (e-mail: xi.lin@xjtu.edu.cn).}
\thanks{X. Tong and M. Yuan are with Huawei Noah's Ark Lab, Hong Kong SAR, China (e-mail: \{tongxialiang, yuan.mingxuan\}@huawei.com).}
\thanks{Q. Zhang is with the Department of Computer Science, City University of Hong Kong, Hong Kong SAR, China (e-mail: qingfu.zhang@cityu.edu.hk).}
}

\markboth{IEEE Transactions on Intelligent Transportation Systems}
{Zhou \MakeLowercase{\textit{et al.}}: (Title)}


\maketitle

\begin{abstract}

In modern intelligent transportation systems (ITS), particularly in freight transportation and logistics, real-time route planning is crucial. It presents unique challenges driven by high uncertainty in service requests, where the number of service customers can vary drastically, ranging from hundreds to thousands. Existing neural methods struggle to maintain performance under such significant variations, which severely limits their practical applicability. To address this crucial shortcoming, this work proposes a novel Instance-Conditioned Adaptation Model (ICAM) designed for better large-scale generalization. In particular, we design a simple yet efficient instance-conditioned adaptation function that adjusts the policy based on the specific geometry and density of the current traffic scenario to improve model adaptability with minimal computational overhead. Furthermore, we propose a powerful yet low-complexity instance-conditioned adaptation module to generate better solutions for instances across various scales. Extensive experiments on synthetic, benchmark, and real-world instances demonstrate that ICAM can consistently achieve promising generalization performance across four widely studied large-scale route planning scenarios. Notably, our proposed method delivers high-quality solutions with remarkably fast inference speed, providing a scalable and efficient solution for real-time intelligent transportation operations. Our code is available at \url{https://github.com/CIAM-Group/ICAM}. 

\end{abstract}


\section{Introduction}
\label{sec:intro}
\IEEEPARstart{A}{s} a fundamental component of route planning, the Vehicle Routing Problem (VRP) plays a crucial role in Intelligent Transportation Systems (ITS), where the solution quality directly affects the transportation cost and service efficiency~\cite{tiwari2023optimization,zhou2024learning_tits,wu2023neural_tits,jia2025arc_tits}, specifically in freight transportation and logistics. The route planning in real-world ITS faces unique challenges driven by the high uncertainty in service requests. A practical ITS algorithm must deliver high-quality performance in real-time across various scenarios.  However, efficiently solving VRPs is a challenging task due to their NP-hard nature. Over the past few decades, extensive metaheuristics, such as LKH3~\cite{LKH3} and HGS~\cite{HGS}, have been proposed to address different VRP variants. In particular, Memetic Algorithms (MAs) have proven to be a representative and powerful class of metaheuristics~\cite{moscato1989MA,zhou2018memetic,zhou2017memetic2}. For example, advanced MAs have been developed to tackle complex VRPs, such as soft-clustered and hard-clustered VRP~\cite{zhou2023bilevel,zhou2025MA_soft_hard}. Although these heuristics have shown promising results for specific problems, their designs heavily rely on expert knowledge and a deep understanding of each problem. Moreover, as the problem scale grows, the search space expands exponentially, making it increasingly challenging for traditional metaheuristics to balance solution quality and computational efficiency. These limitations greatly hinder the practical application of classical heuristic algorithms in time-sensitive ITS applications.

Over the past few years, different neural routing solvers have been explored~\cite{li2022overview,bengio2021machine,zheng2023pareto_tits,li2021heterogeneous_tits,zhang2022learning_tits,ba2026survey}. Compared to leading heuristics, learning-based methods can significantly reduce reliance on handcrafted rules while maintaining competitive solution quality. These neural models learn to solve problems directly from data, offering distinct advantages when expert knowledge is unavailable or expensive to acquire. Among them, constructive neural models incrementally build complete solutions from partial ones for a given instance without iterative search, and they usually have a faster runtime compared to traditional metaheuristics, making them a desirable choice for tackling real-world problems. Existing mainstream constructive neural models can be divided into two categories: supervised learning (SL)-based~\cite{vinyals2015pointer,xiao2023distilling} and reinforcement learning (RL)-based ones~\cite{nazari2018reinforcement,bello2016neural,kool2019attention,zhou2025L2R,zhou2025urs}. The SL-based method requires a large number of problem instances with labels (i.e., near-optimal solutions) as training data. However, obtaining sufficient near-optimal solutions for complex problems is computationally expensive, which impedes practicality. In contrast, RL-based methods can learn neural models by repeatedly interacting with the environment without any labeled data. While current RL-based constructive methods perform well on small-scale instances (e.g., 100-node instances), they struggle to generate reasonably good solutions for instances with much larger scales typical of real-world logistics. This work focuses on label-free RL methods, aiming to address the limitations of large-scale generalization and provide a scalable and efficient solution for real-time intelligent transportation operations.

To address the crucial limitation of RL-based methods on large-scale generalization, two different types of attempts have been explored. The first one is to perform an extra search procedure on model inference to improve the quality of solutions over greedy generation~\cite{hottung2021eas,choo2022sgbs}. However, this approach typically requires expert-designed search strategies and can be time-consuming when dealing with large-scale problems. The second approach is to train the model on instances of varying scales~\cite{khalil2017s2v_dqn,cao2021dan,zhou2023omni}. However, learning cross-scale features effectively for better generalization performance remains a key challenge for RL-based neural routing solvers. 

\begin{figure*}[t]
    \centering
    \begin{subfigure}{0.46\textwidth}
        \centering
        \includegraphics[width=\textwidth]{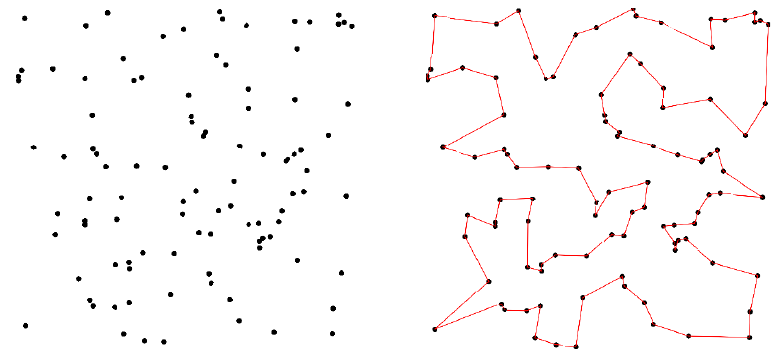} 
        \caption{TSP instance with 100 nodes.}
        \label{subfig:motivation_tsp100}
    \end{subfigure}
    \hfill
    \begin{subfigure}{0.46\textwidth}
        \centering
        \includegraphics[width=\textwidth]{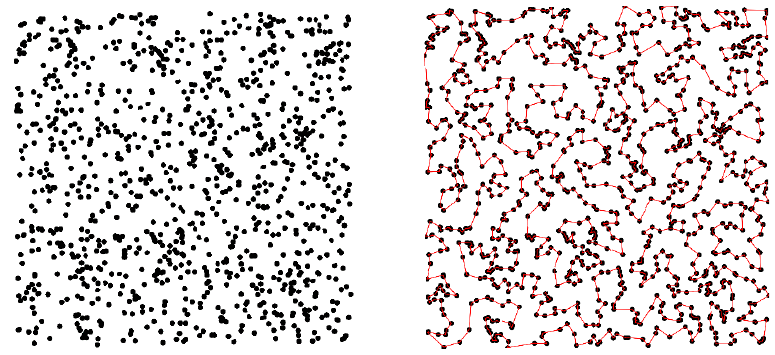}
        \caption{TSP instance with $1,000$ nodes.}
        \label{subfig:motivation_tsp1000}
    \end{subfigure}
    \caption{Comparison of two TSP instances and their optimal solutions with different scales (Left: Instance, Right: Solution). The patterns and geometric structures are quite different for these instances. In this work, we propose a powerful Instance-Conditioned Adaptation Model (ICAM) to leverage these instance-specific patterns to directly generate promising solutions for instances across quite different scales.}
    \label{fig:motivation}
\vskip -0.2in
\end{figure*}

As shown in \cref{fig:motivation}, with different numbers of nodes, the geometric structures of two instances and their optimal solutions are quite different, which could provide valuable information for the solvers. The instance-specific information (e.g., node-to-node distances) has also been leveraged by different RL-based methods, as shown in \cref{table:motivation}. However, regarding the information incorporation strategy, existing methods either simply utilize the node-to-node distances to bias the output score in the decoding phase~\cite{son2023metasage,jin2023pointerformer,wang2024distance} or refine the information via a complex policy~\cite{li2023refiner,gao2023elg}. Although these methods improve the model's convergence efficiency and final performance, they cannot truly capture instance-specific features according to the changes in geometric structures. Therefore, they still struggle to achieve a satisfying generalization performance, especially for large-scale instances.

\begin{table}[t]
\renewcommand\arraystretch{1.1}
\caption{Comparison between our ICAM and existing RL-based neural routing solvers with information incorporation.}
\resizebox{0.48\textwidth}{!}{
\begin{threeparttable}
\begin{tabular}{l|cc|ccc|c}
\bottomrule[0.5mm]
\multirow{2}{*}{Neural Routing Solvers } 
& \multicolumn{2}{c|}{Information}   &  \multicolumn{3}{c|}{Module}  & Varying-scale     \\ 
 & Scale & Distances & Embedding\textdagger &  Attention & Compatibility &Training\\ 
\midrule
S2V-DQN~\cite{khalil2017s2v_dqn}& $\times$   & $\times$  & $\times$                                                           & $\times$   & $\times$  &$\checkmark$ \\
DAN~\cite{cao2021dan}& $\times$   & $\times$  & $\times$                                                            & $\times$   & $\times$  &$\checkmark$ \\
SCA~\cite{kim2022scale}   & $\checkmark$   & $\times$   & $\checkmark$                                                           & $\times$   & $\times$ & $\times$ \\
Meta-AM~\cite{manchanda2022generalization}& $\times$   & $\times$   & $\times$                                                             & $\times$   & $\times$&$\checkmark$ \\
Pointerformer~\cite{jin2023pointerformer}   & $\times$   & $\checkmark$  & $\times$                                                            & $\checkmark$$\ddagger$  & $\times$ & $\times$ \\
Meta-SAGE~\cite{son2023metasage}   & $\checkmark$   & $\checkmark$   & $\checkmark$                                                          & $\times$   & $\checkmark$  & $\times$ \\
FER~\cite{li2023refiner}   & $\times$   & $\checkmark$   & $\checkmark$                                                           & $\times$   & $\times$  &$\times$ \\
Omni\_VRP~\cite{zhou2023omni}& $\times$   & $\times$    & $\times$                                                           & $\times$   & $\times$ &$\checkmark$ \\
ELG~\cite{gao2023elg}   & $\times$   & $\checkmark$    & $\times$                                                         & $\times$   & $\checkmark$ &$\times$ \\
DAR~\cite{wang2024distance}   & $\times$   & $\checkmark$  & $\times$                                                           & $\times$   & $\checkmark$  & $\checkmark$ \\
\midrule
ICAM (Ours)  & $\checkmark$   & $\checkmark$    & $\checkmark$   & $\checkmark$ & $\checkmark$ & $\checkmark$\\

\toprule[0.5mm]
\end{tabular}
\begin{tablenotes}
\large {
\item[ \textdagger] The embedding includes node embedding and context embedding. In FER, the information is used to refine node embeddings via an extra network, and SCA and Meta-SAGE use the scale information to update context embedding. Unlike them, ICAM updates node embeddings by incorporating information into the attention calculations in the encoding phase.

\item[$\ddagger$] In Pointerformer, node-to-node distances are used in the attention calculation of the decoder but are not employed in the encoder.}
\end{tablenotes}
\end{threeparttable}
}
\label{table:motivation}
\vskip -0.2in
\end{table}

In this paper, we propose a powerful RL-based constructive method called \textbf{Instance-Conditioned Adaptation Model (ICAM)}. When facing diverse geometric structures and patterns of instances across different scales, ICAM can effectively capture the instance-specific features (i.e., distance and scale) via the proposed instance-conditioned adaptation function. To make the model better aware of instance-specific information, we incorporate these features into the whole solution construction process (i.e., embedding, attention, and compatibility) via a powerful yet low-complexity instance-conditioned adaptation attention mechanism. Therefore, ICAM can directly generate promising solutions for instances across quite different scales, which improves the large-scale generalization performance for RL-based neural routing solvers. The key contributions of this paper can be summarized as follows:

\begin{itemize}
    \item We design a simple yet efficient instance-conditioned adaptation function to adaptively incorporate the geometric structure of cross-scale instances with a small computational overhead.
    
    \item We systematically investigate the differences in incorporating information between various attention mechanisms, and then further develop a powerful yet low-complexity Adaptation Attention Free Module (AAFM) to explicitly capture instance-specific features during the solution construction process.
    \item Extensive experiments on synthetic, benchmark, and real-world instances demonstrate that ICAM achieves promising generalization performance across four widely studied routing problems (TSP, CVRP, ATSP, and CVRPTW) while maintaining remarkably fast inference speeds.
\end{itemize}

The remainder of this paper is organized as follows. \cref{sec:related_work} reviews related methods for solving routing problems. \cref{sec:icam} presents the details of the proposed ICAM. The experimental results for four different routing problems, along with detailed discussions, are presented in \cref{sec:experiments} and \cref{sec:discussions}, respectively. Furthermore, we provide a comprehensive ablation study in \cref{sec:ablation_study}. Finally, \cref{sec:conclusion} concludes our work and discusses future work.

\section{Related Work}
\label{sec:related_work}

\subsection{VRP in ITS}
The VRP serves as a key problem of modern ITS. Recent progress in ITS has demonstrated promising results of heuristic approaches for complex variants, including classic CVRP~\cite{liu2025vrp_adaptive}, Heterogeneous CVRP~\cite{ba2025multi}, Split and Deliver VRP~\cite{wang2025sdvrp}, and Multi-Depot CVRP~\cite{jiang2025ea_mdvrp}. However, these methods often rely heavily on extensive expert knowledge, which significantly limits their broader applicability. On the other hand, learning-based routing solvers have been increasingly applied to address diverse ITS scenarios, such as CVRP~\cite{li2025vrp_lstm}, Electric CVRP~\cite{lin2025cevrp}, Capacitated Arc Routing~\cite{jia2025arc_tits}, and VRP with Backhauls~\cite{meng2025vrpmblp}. While these contributions have significantly advanced the handling of VRP with specific constraints, the challenge of robust large-scale generalization remains a critical bottleneck for deploying these neural solvers across varying ITS scenarios.

\subsection{Non-conditioned Neural Solvers}
Most neural routing solvers are trained on a fixed scale (e.g., 100 nodes). They usually perform well on instances with the scale trained on, but their performance could drop dramatically on instances with different scales~\cite{kool2019attention,kwon2020pomo}. To mitigate the poor generalization performance, an extra search procedure is usually required to find a better solution. Some widely used search methods include beam search~\cite{joshi2019efficient,choo2022sgbs}, Monte Carlo tree search (MCTS)~\cite{fu2021attgcn_mcts,qiu2022dimes,sun2023difusco}, and active search~\cite{bello2016neural,hottung2021eas}. However, these procedures are very time-consuming, can still perform poorly on instances with significantly different scales, and may require expert-designed strategies for specific problems (e.g., MCTS for TSP). Recently, some methods simplify large-scale VRPs through decomposition policies~\cite{li2021_L2D,pan2023htsp,cheng2023select,ye2023glop,zheng2024udc,li2025drhg,zhou2025dualopt}. Although these methods possess good generalization abilities, they typically rely on well-designed divide-and-conquer policies and overlook the dependency between stages, which makes model design challenging and heavily dependent on expert knowledge. In contrast, our constructive solver relies minimally on domain knowledge.

\subsection{Varying-scale Training in Neural Solvers}
Training the neural solver directly on instances with different scales is another popular method to improve its generalization performance. Expanding the training scale can bring a broader range of cross-scale data. Training using these data enables the model to learn more scale-independent features, thereby achieving better large-scale generalization performance. This straightforward approach can be traced back to \cite{vinyals2015pointer} and \cite{khalil2017s2v_dqn}, which try to train the model on instances with varying scales to improve solving performance. Subsequently, a series of works have been developed to utilize the varying-scale training scheme to improve the generalization performance of their own neural solvers~\cite {lisicki2020evaluating,cao2021dan,manchanda2022generalization,zhou2023omni,wang2024distance}. However, current RL-based models also face the challenge of efficiently capturing cross-scale features from varying-scale training data, which severely hinders their generalization ability on large-scale problems. Similar to the varying-scale training scheme, a few SL-based neural routing solvers learn to construct partial solutions with various scales during training and achieve a good generalization performance~\cite{luo2023lehd,drakulic2023bq}. Nevertheless, in real-world applications, it could be very difficult to obtain high-quality labeled solutions for SL-based model training. To address this, some recent works utilize model-based search techniques to generate high-quality pseudo-labels~\cite{luo2025Boosting}. While this realizes effective SL-like training, designing advanced search strategies often requires significant domain knowledge, which introduces an additional computational cost to the training process. In contrast, our RL-based method learns directly from interactions without external labels and effectively captures cross-scale features, enabling ICAM to achieve promising generalization on instances of varying scales.

\subsection{Information-conditioned Neural Solvers}
Recently, several studies have shown that incorporating auxiliary information (e.g., the distance between each pair of nodes for VRPs) can facilitate model training and enhance solving performance. In \cite{kim2022scale}, the scale-related feature is added to the context embedding of the decoder to make the model scale-aware during the decoding phase. \cite{jin2023pointerformer}, \cite{son2023metasage} and \cite{wang2024distance} use the distance to bias the output score in the decoding phase, thereby guiding the model toward more efficient exploration. Especially, \cite{gao2023elg} employ a local policy network to catch distance knowledge and integrate it into the compatibility calculation, and in \cite{li2023refiner}, the distance-related feature is utilized to refine node embeddings to improve the model exploration. None of them incorporates the information into the whole neural solution construction process and fails to achieve satisfactory generalization performance on large-scale instances. Other recent studies employ distance-based search space reduction strategies~\cite{xiao2025dgl,chen2025ttpl}. While reducing the computational difficulty via distance-based search space reduction, manually restricting the search space increases the risk of the model becoming trapped in local optima. Our method avoids this by retaining global awareness via soft attention biases.

By systematically analyzing the existing works, we find that the following two aspects are very important for neural solvers to obtain better generalization performance: 
\begin{itemize}
    \item \textbf{Instance-Conditioned Information:} Given that routing instances exhibit diverse geometric structures and patterns across different scales, effectively capturing instance-specific features is crucial for achieving robust generalization performance.
    \item \textbf{Varying-Scale Training:} Training the neural solver on instances of varying scales facilitates the learning of more scale-independent features, thereby improving large-scale generalization performance.
\end{itemize}

Unlike many existing baselines, ICAM effectively leverages the synergy between instance-conditioned information and varying-scale training. It achieves superior large-scale generalization performance through a streamlined, constructive architecture. ICAM does not rely on complex multi-stage designs, expensive labeled data, or risk-prone search space restrictions, while still maintaining fast inference runtime to construct high-quality solutions. These characteristics meet critical and urgent requirements in freight transportation and logistics.

\section{Instance-Conditioned Adaptation}
\label{sec:icam}

In this section, we detail our proposed ICAM for large-scale generalization. The core of ICAM is a novel instance-conditioned adaptation function, $f(N,d_{ij})$, designed to capture instance-specific scale and distance information. This function is deeply integrated into two key components of ICAM: 1) our proposed efficient Adaptation Attention Free Module (AAFM), which replaces traditional MHA, and 2) the compatibility calculation, which further enhances the solution quality by combining the model's learned selection bias with the raw geometric bias of instances. Finally, we employ a three-stage varying-scale training strategy for the model. The following subsections will describe the design specifics of each component in sequence.

\subsection{Instance-Conditioned Adaptation Function}
\label{sec:icaf}
In this work, we propose a straightforward yet efficient instance-conditioned adaptation function $f(N,d_{ij})$ to incorporate the instance-specific information into the neural model:
\begin{equation}
f(N,d_{ij}) = - \alpha \cdot \log _2N \cdot d_{ij} \quad \forall i,j \in 1, \ldots, N,
\label{eq:adaptation_function}
\end{equation}
    
where $N$ is the scale information (e.g., the total number of nodes), $d_{ij}$ represents the distance between node $i$ and node $j$, and $\alpha > 0$ is the learnable parameter. From an ITS perspective, $f(N, d_{ij})$ serves as a dynamic traffic state adapter. Specifically, $N$ represents the fluctuation in real-world traffic demand (e.g., peak vs. off-peak periods), while $d_{ij}$ encodes the spatial distribution of service requests. This explicit conditioning allows ICAM to adaptively adjust its exploration strategy in response to varying traffic states. For the proposed $f(N,d_{ij})$, to enable the model to better perceive the underlying structural changes across varying-scale instances, we perform the following operations:

\begin{itemize}
    \item The pair-wise distance $d_{ij}$ is the most direct instance-specific feature in routing problems. Our adaptation function is designed to yield a larger score for shorter distances, which is achieved by the negative sign in $f(N,d_{ij})$ and given adaptation weight $\alpha > 0$.
    
    \item The scale $N$ is another crucial instance-specific feature as it directly reflects the node density. We use its logarithm (i.e., $\log_2N$) to enable the model to effectively perceive density changes and avoid extreme numerical values in large-scale instances.

    \item This only learnable parameter $\alpha$ allows the model to automatically learn the degree of adaptability across varying-scale instances with minimal extra time and memory overhead, enabling the model to maintain high efficiency when tackling large-scale instances.
\end{itemize}

By incorporating $f(N,d_{ij})$ throughout the entire neural solution construction process (AAFM and Compatibility), the model is expected to be more aware of instance-specific information and, consequently, generate a better solution for each instance.

\paragraph{Less Is More}
To demonstrate the superiority of our proposed function $f(N,d_{ij})$, we report its performance on solving the TSP1000 instances using the seminal POMO model~\cite{kwon2020pomo}, and compare it with three typical information incorporation approaches: (1) Simple node-to-node distances~\cite{jin2023pointerformer,wang2024distance}; (2) Node-to-node distances with a bias coefficient $\alpha$ introduced~\cite{son2023metasage}; and (3) An extra local policy as adopted in \cite{gao2023elg}. We incorporate four different incorporation approaches into all attention calculations (see \cref{eq:mha_adaptation}) in both the encoder and decoder, respectively. Moreover, we also combine them with the compatibility calculation in the decoder~\cite{gao2023elg,wang2024distance}. Note that we train all models solely on 100-node instances with $100$ epochs. At each epoch, we process $1,000$ batches of instances, each with a batch size of $bs = 64$, for all models. The remaining model parameters are consistent with those of the original POMO model.

\begin{table}[t]
\centering
\caption{Comparison on TSP1000 instances with different instance-specific information incorporation approaches.}
\resizebox{0.5\textwidth}{!}{
\begin{tabular}{ll | c |ccc }
\toprule[0.5mm]
\multicolumn{2}{l|}{Method}  & Params & Avg.memory & Gap & Time \\
\midrule

\multicolumn{2}{l|}{POMO} &1.27M & 107.50MB & 25.916\% &63.80s  \\
\multicolumn{2}{l|}{POMO + dist.} &1.27M & 124.22MB & 22.696\% & 83.85s \\
\multicolumn{2}{l|}{POMO + $\alpha$ * dist.} &1.27M & 124.22MB & 14.517\% &86.23s  \\
\multicolumn{2}{l|}{POMO + Local policy} &1.30M & 163.44MB & 14.821\% & 130.26s \\
\multicolumn{2}{l|}{POMO + $f(N,d_{ij})$} &1.27M & 124.22MB & \cellcolor[HTML]{D0CECE}\textbf{10.812\%} & 86.92s \\

\bottomrule[0.5mm]
\end{tabular}
}
\label{table:incorporation_approaches}
\vskip -0.2in
\end{table}
As shown in \cref{table:incorporation_approaches}, these impressive results highlight the effectiveness of our proposed function $f(N,d_{ij})$. Compared with other approaches, our proposed function can significantly improve the generalization of the original model with a very small time and memory overhead.

\subsection{Instance-Conditioned Adaptation Model}
\label{sec:detailed_icam}

In addition to the instance-conditioned adaptation function, the model structure is also crucial to achieve a promising generalization performance. Most existing models adopt the encoder-decoder structure, which is developed from Transformer~\cite{kool2019attention,gao2023elg}. Without loss of generality, taking well-known POMO~\cite{kwon2020pomo} as an example, this subsection briefly reviews the prevailing neural solution construction pipeline and discusses how to efficiently incorporate the instance-specific information.

\paragraph{Rethinking Attention Mechanism in Neural Routing Solvers}

Given an instance $S=\{\mathbf{s}_i\}_{i=1}^N$, $\mathbf{s}_i$ represents the features of each node (e.g., the coordinates of each city in TSPs). These features are transformed into initial embeddings $H^{(0)}=(\mathbf{h}_1^{(0)},\ldots,\mathbf{h}_N^{(0)})$ via a linear projection. The initial embeddings pass through $L$ attention layers to get node embeddings $H^{(L)}=(\mathbf{h}_1^{(L)},\ldots,\mathbf{h}_N^{(L)})$. The attention layer consists of a Multi-Head Attention (MHA) sub-layer~\cite{vaswani2017attention} and a Feed-Forward (FF) sub-layer. During the decoding process, POMO model generates a solution in an autoregressive manner. For the example of TSP, in the $t$-step construction, the context embedding is composed of the first visited node embedding and the last visited node embedding, i.e., $\mathbf{h}_{(C)}^{t}=[\mathbf{h}_{\pi_1}^{(L)}, \mathbf{h}_{\pi_{t-1}}^{(L)}]$. The new context embedding $\hat{\mathbf{h}}_{(C)}^{t}$ is then obtained via the MHA operation on $\mathbf{h}_{(C)}^{t}$ and $H^{(L)}$. Finally, the model yields the selection probability for each unvisited node $p_{\boldsymbol{\theta}}(\pi_t = i \mid S , \pi_{1:t-1})$ by calculating compatibility on $\hat{\mathbf{h}}_{(C)}^{t}$ and $H^{(L)}$.

\begin{figure*}[t]
\begin{center}
\centerline{\includegraphics[width=\textwidth]{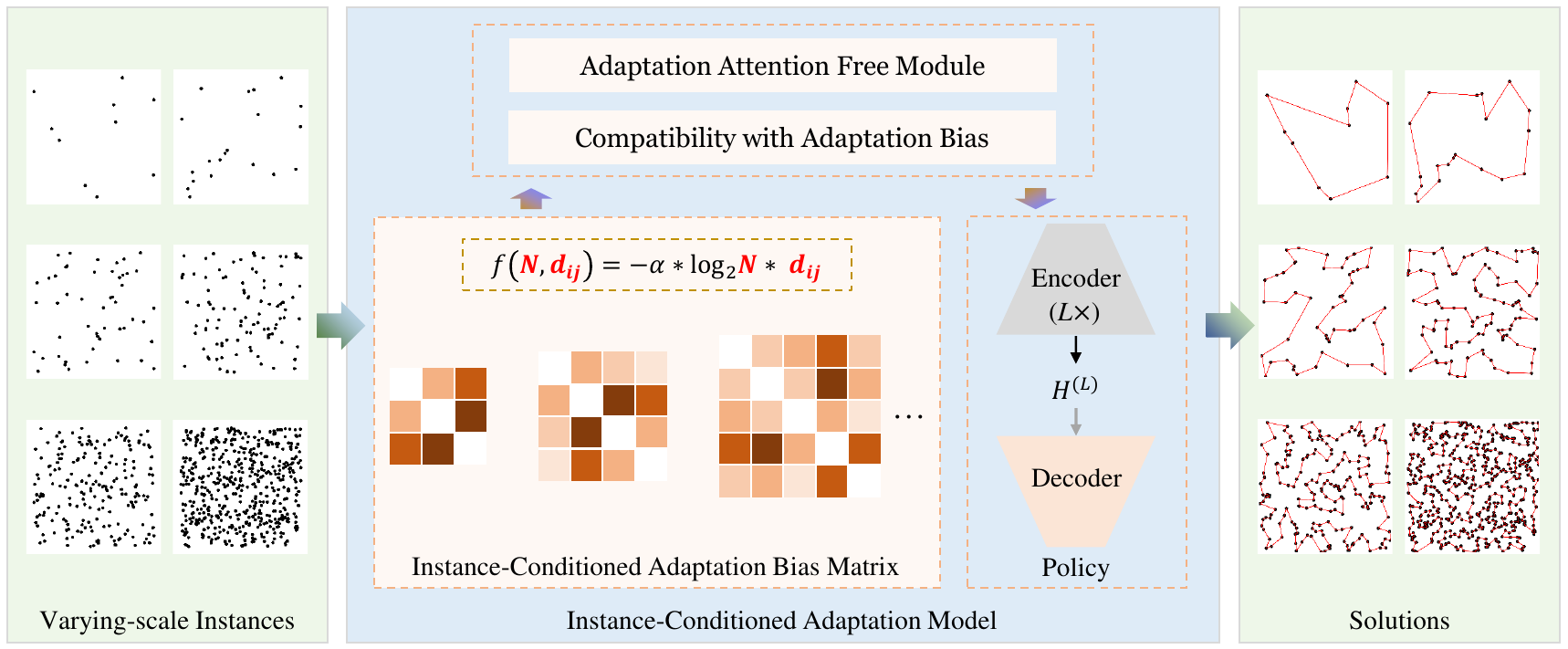}}
\caption{The proposed ICAM. Taking the TSP as an example, comprehensive instance-conditioned information is incorporated into the whole solution construction process. ICAM solves the specific instance by adaptively updating the corresponding adaptation bias matrix. Specifically, we utilize AAFM to replace all MHA operations and combine $f(N,d_{ij})$ with the compatibility calculation.}
\label{fig:ICAM}
\end{center}
\vskip -0.4in
\end{figure*}

From the above description, MHA operation is the core component of Transformer-like routing solvers. In the self-attention mode, MHA performs a scaled dot-product attention for each head. The self-attention calculation is written as
\begin{equation}
Q = XW^{Q}, \quad K = XW^{K}, \quad V = XW^{V},
\label{eq:qkv}
\end{equation}
\begin{equation}
    \mathrm{Attention}(Q, K, V) = \mathrm{softmax}\left(\frac{QK^{\mathrm{T}}}{\sqrt{d_k}}\right)V,
    \label{eq:mha}
\end{equation}
where $X$ represents the input, $W^{Q}$, $W^{K}$ and $W^{V}$ are three learning matrices, and $d_k$ is the dimension for $K$. In a Transformer-based routing solver, the MHA incurs primary memory usage and computational cost. Additionally, the MHA calculation is not suitable for capturing the relationship between nodes. It cannot directly take advantage of the pair-wise distances between nodes.

\begin{figure}[t]
\begin{center}
\centerline{\includegraphics[width=0.87\columnwidth]{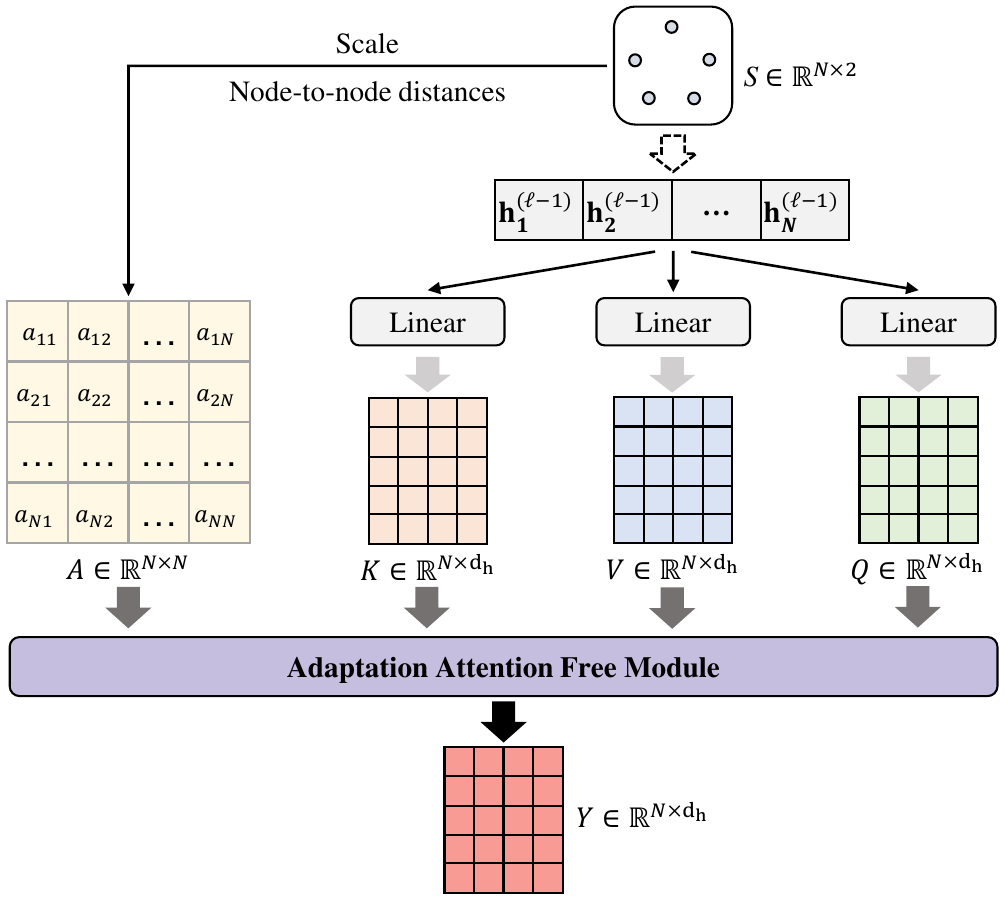}}
\caption{The Structure of AAFM.}
\label{fig:aafm_structure}
\end{center}
\vskip -0.3in
\end{figure}

\begin{figure}[t]
\begin{center}
\centerline{\includegraphics[width=0.7\columnwidth]{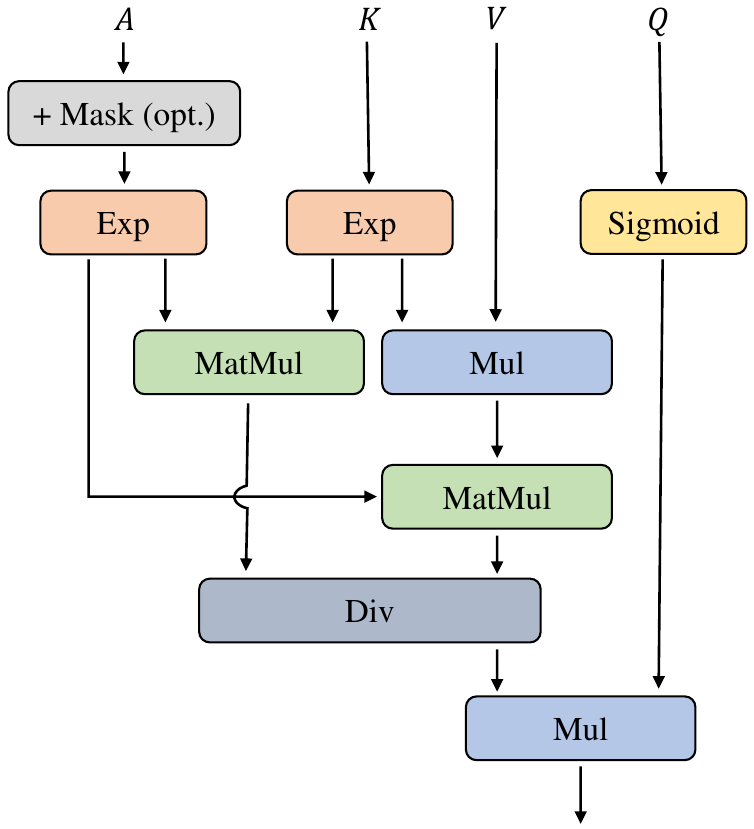}}
\caption{The Detailed Calculation Process of AAFM. }
\label{fig:aafm_detailed}
\end{center}
\vskip -0.3in
\end{figure}

\paragraph{Adaptation Attention Free Module} 
As shown in \cref{fig:ICAM}, the proposed ICAM is also developed from the encoder-decoder structure, we remove all high-complexity MHA operations in both the encoder and decoder, and replace them with the proposed novel module, named \textbf{A}daptation \textbf{A}ttention \textbf{F}ree \textbf{M}odule (AAFM). As shown in \cref{fig:aafm_structure}, AAFM is based on the AFT-full operation~\cite{zhai2021aft}, which offers more excellent speed and memory efficiency than MHA. The proposed AAFM can be expressed as
\begin{equation}
\mathrm{AAFM}(Q,K,V,A) = \sigma (Q) \odot \frac{\exp (A) (\exp (K) \odot V)}{\exp (A)\exp (K)},
\label{eq:AAFM}
\end{equation}
where $Q,K,V$ are also separately obtained via \cref{eq:qkv}, $\sigma$ represents Sigmoid function, $\odot$ represents the element-wise product, $A=\{\mathbf{a}_{ij}\}$, $\forall i,j \in 1, \ldots, N$ denotes the pair-wise adaptation bias computed by our adaptation function $f(N,d_{ij})$ in \cref{eq:adaptation_function}. The detailed calculation of AAFM is shown in \cref{fig:aafm_detailed}. 

We motivate this design choice by highlighting the limitations of the standard MHA, namely its inability to explicitly learn relative position biases and its high computational complexity, especially for large-scale instances. In contrast, the proposed AAFM can explicitly capture relative position biases via the designed adaptation function $f(N,d_{ij})$ (as the bias matrix $A$). By injecting the scale and distance information into AAFM, it can generate more informative node representations $H^{(L)}$ and current state $\hat{\mathbf{h}}_{(C)}^{t}$, which inherently reflect the unique geometric structure of each instance (as shown in \cref{fig:tsp_hm_short}). Furthermore, AAFM exhibits lower computational overhead than MHA, resulting in lower complexity and a faster inference time. For a detailed discussion of the comparison of AAFM and MHA, please refer to \cref{sec:aft_mha}.

\paragraph{Compatibility with Adaptation Bias}
To further improve the solving performance, we integrate $f(N,d_{ij})$ into the compatibility calculation \cite{son2023metasage,gao2023elg}. The new compatibility $u^{t}_{i}$ can be expressed as 
\begin{equation}
    u^{t}_{i} = 
    \begin{cases}
    \xi  \cdot \text{tanh}(\frac{\hat{\mathbf{h}}_{(C)}^{t}(\mathbf{h}_{i}^{(L)})^\mathrm{T}}{\sqrt{d_k}} +a_{t-1,i})   & \text{if} \ i \not\in \{\pi_{1:t-1}\} \\
    -\infty  &\text{otherwise}
\end{cases},
    \label{eq:AFT_u_calculated}
\end{equation}
\begin{equation}
    p_{\boldsymbol{\theta}}(\pi_t = i \mid S , \pi_{1:t-1}) = \frac{e^{u_{i}^{t}}}{\sum_{j=1}^{N}e^{u_{j}^{t}}},
    \label{eq:softmax}
\end{equation}
where $\xi$ is the clipping parameter, $\hat{\mathbf{h}}_{(C)}^{t}$ and $\mathbf{h}_{i}^{(L)}$ are calculated via AAFM instead of MHA. $a_{t-1,i}$ represents the adaptation bias between each remaining node and the current node. Note that in the AAFM of the decoder, the node masking state in the current step $t$ is added to $A$ additionally.

By incorporating the adaptation function $a_{t-1,i}$ into the compatibility calculation, we provide effective guidance for making final decisions. This design allows the policy to fully combine two distinct sources of bias: (1) the learned selection bias from the model itself, and (2) the explicit geometric bias $a_{t-1,i}$ provided by the instance itself, which indicates spatially favorable nodes. Through the proposed compatibility calculation with adaptation bias, ICAM can make more informative node selections, which is especially critical for large-scale instances.

This dual-integration strategy is validated by our ablation study, as shown in Appendix \ref{append:ablation_study}. The results clearly show that applying the function in both modules (AAFM+CAB) achieves significantly better performance than applying it in only one. All learnable biases $\alpha$ are simply trained end‑to‑end together with all other model parameters. We provide the convergence curves of learnable bias $\alpha$ throughout the training process for different problems in Appendix \ref{append:alpha_change}.
 
Finally, the probability of generating a complete solution $\mathbf{\pi}$ for instance $X$ is calculated as
\begin{equation}
    p_{\boldsymbol{\theta}}(\mathbf{\pi}\mid S)=\prod_{t=2}^{N}
    p_{\boldsymbol{\theta}}(\pi_t\mid S,\pi_{1:t-1}).
    \label{eq:construct_solution}
\end{equation}

\subsection{Varying-scale Training}

To enable the model to be aware of the scale information better and simultaneously learn various pair-wise biases of training instances at different scales, we develop a three-stage training scheme to enable the proposed ICAM to incorporate instance-conditioned information more effectively. Note that to ensure a fair comparison with baseline models, we additionally conduct an ablation study about comparison under consistent varying-scale training settings. For experimental results and analysis, please refer to Appendix \ref{ablation:vst_comparison}. The detailed settings of our three-stage training scheme are as follows:
\paragraph{Stage 1: Warming-up on Small-scale Instances} We employ a warm-up procedure in the first stage. Initially, the model is trained for several epochs on small-scale instances. For example, we use a total of $256,000$ randomly generated TSP100 instances for each epoch in the first stage to train the model for 100 epochs. A warm-up training can make the model more stable in the subsequent varying-scale training.

\paragraph{Stage 2: Learning on Varying-scale Instances} In the second stage, we train the model on varying-scale instances for much longer epochs. For each batch, the scale $N$ is randomly sampled from the discrete uniform distribution \textbf{Unif}([$100$,$500$]) for all problems. Considering GPU memory constraints, we decrease the batch size with the scale increases. ICAM is trained by the REINFORCE~\cite{williams1992reinforce} algorithm, and the loss function (denoted as $\mathcal{L}_{\mathrm{POMO}}$) used in the first and second stages is the same as in POMO~\cite{kwon2020pomo}. The gradient ascent with an approximation of the loss function can be written as
\begin{equation}
    \nabla_\theta \mathcal{L}_{\mathrm{POMO}} (\theta) \approx \frac{1}{BN} \sum_{m=1}^B\sum_{i=1}^N  
    G_{m,i}
    \nabla_\theta \log p_\theta\left(\pi^i \mid S_{m}\right),
    \label{eq:mean_loss}
\end{equation}
\begin{equation}
    G_{m,i}= R\left(\pi^{i}\mid S_{m} \right)-b^i(S_{m} ),
    \label{eq:advantage}
\end{equation}
\begin{equation}
    b^i(S_{m})=\frac{1}{N} \sum_{j=1}^N R\left(\pi^j\mid S_{m}\right) \quad \text { for all } i .
    \label{eq:shared_baseline}
\end{equation}
where $R\left(\pi^{i}\mid S_{m} \right)$ represents the total reward (e.g., the negative value of tour length) of instance $S_{m}$ given a specific solution $\pi^{i}$. \cref{eq:shared_baseline} is a baseline as adopted in \cite{kwon2020pomo}. 
\paragraph{Stage 3: Top-$k$ Elite Training} Under the POMO structure, $N$ trajectories are constructed in parallel for each instance during training. In the third stage, we want the model to focus more on the best $k$ trajectories among all $N$ trajectories. To achieve this, we design a new loss $\mathcal{L}_{\mathrm{Top}}$, $\mathcal{L}_{\mathrm{Top}}$ only focus on the $k$ best trajectories out of $N$ trajectories, and its gradient ascent can be expressed as
\begin{equation}
    \nabla_\theta \mathcal{L}_{\mathrm{Top}} (\theta) \approx \frac{1}{Bk} \sum_{m=1}^B\sum_{i=1}^k 
    G_{m,i} \nabla_\theta \log p_\theta\left(\pi^i \mid S_{m}\right).
    \label{eq:topk_loss}
\end{equation}
We combine $\mathcal{L}_{\mathrm{Top}}$ with $\mathcal{L}_{\mathrm{POMO}}$ as the joint loss in the training of the third stage, i.e., 
\begin{equation}
    \mathcal{L}_{\mathrm{Joint}} = \mathcal{L}_{\mathrm{POMO}} + \beta  \cdot \mathcal{L}_{\mathrm{Top}}.
    \label{eq:double_loss}
\end{equation}
where $\beta \in [0,1]$ is a coefficient balancing the original loss and the new loss.

\section{Experiments}
\label{sec:experiments}
In this section, we conduct a comprehensive comparison between ICAM and other classical and learning-based solvers using Traveling Salesman Problem (TSP), Capacitated Vehicle Routing Problem (CVRP), Asymmetric Traveling Salesman Problem (ATSP), and CVRP with Time Windows (CVRPTW) instances of different scales.

\subsection{Relevance to Intelligent Transportation Systems}
The four VRP variants addressed in this study (TSP, CVRP, ATSP, and CVRPTW) serve as representative case studies within the ITS context, which represent diverse contingencies and emergency situations encountered in real-world transportation scenarios. Specifically, capacity limits and time windows simulate delivery fleets operating under finite cargo space and strict service-level agreements. Furthermore, the cost structures in TSP and ATSP capture varying urban topologies, where asymmetric costs reflect real-world restrictions such as one-way streets or unexpected road closures.

\subsection{Experimental Setup}
\paragraph{Problem Setting}
For all problems, the instances of training and testing are generated randomly. Specifically, we generate the instances with a setup as prescribed in~\cite{kool2019attention} for TSPs and CVRPs, and we follow the data generation method in MatNet~\cite{kwon2021matnet} for ATSP. For the test set, unless stated otherwise, we generate $10,000$ instances for the $100$-node case and $128$ instances for cases with larger scale. Specifically, for capacity settings in CVRP and CVRPTW, we follow the approach in~\cite{luo2023lehd,drakulic2023bq}. For CVRPTW, we use the problem setup provided by \cite{zhou2024mvmoe}.

\paragraph{Model Architecture}
\label{sec:model_structure}
Our proposed function $f(N,d_{ij})$ and AAFM are adaptable to different models, depending on the specific problem. For TSPs, CVRPs, and CVRPTW, ICAM is developed from the well-known POMO model~\cite{kwon2020pomo}. For CVRPTW, we make only minimal adjustments to our CVRP model for incorporating additional time-related information, which includes: (1) we augment the node input features to include the time window (i.e., earliest and latest arrival times) in addition to coordinates and demand, and (2) we add the end time of serving the current node to the decoder's context embedding to account for the time constraint. Considering the specificity of ATSPs, we replace the backbone network with MatNet~\cite{kwon2021matnet}. We remove all original attention calculations in POMO and MatNet (including both the encoder and decoder) and replace them with our proposed AAFM. Additionally, as shown in \cref{eq:AFT_u_calculated}, in the decoding phase, we modify the compatibility calculation by adding our adaptation function $f(N,d_{ij})$, following the approach in ~\cite{gao2023elg} and \cite{son2023metasage}, so as to improve the model performance further. Notably, we retain our original adaptation function ($f(N, d_{ij})$) to all tested problems, which only incorporates distance and scale information. We observe that this function already provides a robust foundation for generalization, enabling the model to effectively handle large-scale CVRPTW instances.

\begin{table}[h!]
\centering
\caption{Model and training settings in our experiments.}
\label{table:hyperparameters}
\resizebox{0.49\textwidth}{!}{
\begin{tabular}{ll c c c c}
\toprule[0.5mm]
\multicolumn{2}{l}{ }& \textbf{TSP}   & \textbf{CVRP} & \textbf{ATSP} & \textbf{CVRPTW}\\  
\midrule
\multicolumn{2}{l}{Optimizer}              & \multicolumn{4}{c}{Adam}       \\
\multicolumn{2}{l}{Clipping parameter}              & \multicolumn{4}{c}{50} \\
\multicolumn{2}{l}{Initial learning rate}              & \multicolumn{4}{c}{$10^{-4}$} \\
\multicolumn{2}{l}{Learning rate of stage 3}              & \multicolumn{4}{c}{$10^{-5}$} \\
\multicolumn{2}{l}{Initial $\alpha$ value}              & \multicolumn{4}{c}{1} \\
\multicolumn{2}{l}{Loss function of stage 1 \& 2}  & \multicolumn{4}{c}{$\mathcal{L}_{\mathrm{POMO}}$} \\
\multicolumn{2}{l}{Loss function of stage 3}              & \multicolumn{4}{c}{$\mathcal{L}_{\mathrm{Joint}}$} \\
\multicolumn{2}{l}{Parameter $\beta$ of stage 3}  & \multicolumn{4}{c}{$0.1$} \\
\multicolumn{2}{l}{Parameter $k$ of stage 3} & \multicolumn{4}{c}{$20$} \\
\multicolumn{2}{l}{The number of encoder layer} & \multicolumn{4}{c}{$12$}            \\
\multicolumn{2}{l}{Embedding dimension}       & \multicolumn{4}{c}{$128$}          \\
\multicolumn{2}{l}{Feed forward dimension}    & \multicolumn{4}{c}{$512$}         \\
\multicolumn{2}{l}{Batches of each epoch}      & \multicolumn{4}{c}{$1,000$}         \\
\multicolumn{2}{l}{Scale of stage 1}         &\multicolumn{4}{c}{$100$}       \\
\multicolumn{2}{l}{Scale of stage 2 \& 3}         &\multicolumn{4}{c}{$[100,500]$}      \\
\multicolumn{2}{l}{Epochs of stage 1}         &\multicolumn{4}{c}{$100$}       \\
\multicolumn{2}{l}{Epochs of stage 3}         & \multicolumn{4}{c}{$200$}       \\
\multicolumn{2}{l}{Epochs of stage 2}         & $2,200$        & $700$  & $100$ & $700$     \\
\multicolumn{2}{l}{Capacity of stage 1}         & $-$           & $50$  & $-$  & $50$   \\
\multicolumn{2}{l}{Capacity of stage 2 \& 3}         & $-$           & $[50,100]$  & $-$  & $[50,100]$    \\
\multicolumn{2}{l}{Batch size of stage 1}      & $256$           & $128$    & $256$  & $128$        \\
\multicolumn{2}{l}{Batch size of stage 2 \& 3}      & $\left [ 160 \times (\frac{100}{N} )^2 \right ] $           & $\left [ 128 \times (\frac{100}{N} )^2 \right ] $ & $\left [ 160 \times (\frac{100}{N} )^2 \right ] $   & $\left [ 128 \times (\frac{100}{N} )^2 \right ] $  \\
\multicolumn{2}{l}{Gradient clipping}      & $-$           & max\_norm=$5$    & max\_norm=$5$    & $-$     \\
\multicolumn{2}{l}{Weight decay}                &$-$ &$-$& $10^{-6}$  &$-$  \\
\multicolumn{2}{l}{Total epochs}         & $2,500$       & $1,000$    & $400$    & $1000$ \\
\bottomrule[0.5mm]
\end{tabular}
}
\vspace{-10pt}
\end{table}
\paragraph{Model Setting} 
For all experiments, the embedding dimension is set to $128$, and the dimension of the feed-forward layer is set to $512$. We set the number of attention layers in the encoder to $12$ to generate better node embeddings\footnote{For ATSP model, the 12-layer encoder represents two independent 6-layer encoders, following MatNet architecture~\cite{kwon2021matnet}.}. For the ATSP model, we change the dimension of the input feature to $50$, i.e., the distance of the 50 nearest nodes to each node in both row and column elements. Further, these features are transformed into different initial embeddings via different 128-dimensional linear projections in a 6-layer row encoder and a 6-layer column encoder, respectively. The clipping parameter $\xi  = 50$ in \cref{eq:AFT_u_calculated} for better training convergence~\cite{jin2023pointerformer}. We train and test all experiments using a single NVIDIA GeForce RTX 3090 GPU with 24GB of memory.

\paragraph{Training Setting}
For all models, we use Adam~\cite{kingma2014adam} as the optimizer, and the initial learning rate $\eta $ is set to $10^{-4}$. Every epoch, we process $1,000$ batches for all problems. For each instance, $N$ different solutions are always generated in parallel, following in ~\cite{kwon2020pomo}. The rest of the training settings are as follows:
\begin{enumerate}
    \item In the first stage of the process, we set different batch sizes for different problems due to memory constraints: $256$ for (A)TSP and $128$ for CVRP(TW). We use instances for a scale of $100$ to train corresponding models for $100$ epochs. Additionally, the capacity for CVRP(TW) instances is fixed at $50$. 
    \item In the second stage, we optimize memory usage by adjusting batch sizes according to the changed scale. For (A)TSP, the batch size $bs=\left [ 160 \times (\frac{100}{N} )^2 \right ] $. In the case of CVRP(TW),  the batch size $bs=\left [ 128 \times (\frac{100}{N} )^2 \right ] $. We train the TSP model for $2,200$ epochs and the CVRP(TW) model for $700$ epochs in this stage. For the ATSP model, the training duration is $100$ epochs attributed to the fast convergence. Furthermore, the capacity of each batch is consistently set by randomly sampling from \textbf{Unif}([$50$,$100$]) for CVRP(TW). 
    \item In the last stage, we adjust the learning rate $\eta $ to $10^{-5}$ across all models to enhance model convergence and training stability. The parameter $\beta$ and $k$ are set to $0.1$ and $20$, respectively. The training period is standardized to $200$ epochs for all models, and other settings are consistent with the second stage. 
\end{enumerate}
Note that for each problem, we use the same model on all scales and distributions. The detailed information about the hyperparameter settings can be found in \cref{table:hyperparameters}.

\begin{table*}[t]
\centering
\caption{Comparison on routing problems (TSP, CVRP, ATSP, and CVRPTW) with uniform distribution and scale $\leq 1,000$. }
\resizebox{0.98\textwidth}{!}{
\begin{tabular}{ll | ccc | ccc |  ccc | ccc }
\toprule[0.5mm]

\multicolumn{2}{l|}{ }& \multicolumn{3}{c|}{TSP100 }  & \multicolumn{3}{c|}{TSP200}  & \multicolumn{3}{c|}{TSP500} & \multicolumn{3}{c}{TSP1000}\\ 
\multicolumn{2}{l|}{Method}  & Obj. & Gap & Time & Obj. & Gap & Time & Obj. & Gap & Time  & Obj. & Gap & Time\\
\midrule
\multicolumn{2}{l|}{LKH3}   &7.7632 & 0.000\% & 56m & 10.7036 & 0.000\% & 4m & 16.5215 & 0.000\% & 32m & 23.1199 & 0.000\% & 8.2h \\
\multicolumn{2}{l|}{Concorde} &7.7632 & 0.000\% & 34m & 10.7036 & 0.000\% & 3m & 16.5215 & 0.000\% & 32m & 23.1199 & 0.000\% & 7.8h \\

\midrule
\multicolumn{2}{l|}{Att-GCN+MCTS*}  & \cellcolor[HTML]{D0CECE}\textbf{7.7638} & \cellcolor[HTML]{D0CECE}\textbf{0.037\%} & 15m & 10.8139 & 0.884\% & 2m & 16.9655 & 2.537\% & 6m & 23.8634 & 3.224\% & 13m\\
\multicolumn{2}{l|}{DIMES AS+MCTS*}   & $-$  & $-$ & $-$ & $-$& $-$ & $-$& 16.84 & 1.76\% & 2.15h & 23.69 & 2.46\% &4.62h\\
\multicolumn{2}{l|}{SO-mixed*}   & $-$ & $-$ & $-$ &10.7873 & 0.636\% & 21.3m& 16.9431 & 2.401\% & 32m & 23.7656 & 2.800\% & 55.5m\\
\multicolumn{2}{l|}{DIFUSCO greedy+2-opt*} & 7.78 & 0.24\% & $-$ & $-$& $-$ & $-$& 16.80 & 1.49\% & 3.65m & 23.56 & 1.90\% &12.06m\\
\multicolumn{2}{l|}{T2T sampling*} & $-$  & $-$ & $-$ & $-$& $-$ & $-$& 17.02 & 2.84\% & 15.98m & 24.72 & 6.92\% &53.92m\\

\multicolumn{2}{l|}{H-TSP} & $-$ & $-$ & $-$ & $-$& $-$ & $-$& 17.549 & 6.220\% &23s  & 24.7180 & 6.912\% &47s\\
\multicolumn{2}{l|}{GLOP (more revisions)}   & 7.7668 & 0.046\% & 1.9h & 10.7735& 0.653\% & 42s& 16.8826 & 2.186\% & 1.6m &23.8403  & 3.116\% &3.3m\\

\midrule
\multicolumn{2}{l|}{BQ greedy}  & 7.7903 & 0.349\% & 1.8m &10.7644 & 0.568\% & 9s& 16.7165 & 1.180\% &  46s& 23.6452 & 2.272\% &1.9m\\
\multicolumn{2}{l|}{LEHD greedy}    & 7.8080 & 0.577\% & 27s & 10.7956 & 0.859\% & 2s & 16.7792 & 1.560\% & 16s & 23.8523 & 3.168\% & 1.6m\\

\midrule
\multicolumn{2}{l|}{POMO aug$\times$8}    & 7.7736 & 0.134\% & 1m & 10.8677&1.534\%  & 5s& 20.1871 & 22.187\% & 1.1m & 32.4997 & 40.570\% & 8.5m\\
\multicolumn{2}{l|}{ELG aug$\times$8}  &7.7807 & 0.225\% &3m  &10.8620 & 1.480\% &13s & 17.6544 & 6.857\% & 2.3m &25.5769 &10.627\% & 15.4m\\

\multicolumn{2}{l|}{Pointerformer aug$\times$8}  & 7.7759 & 0.163\% & 49s &10.7796 & 0.710\% & 11s & 17.0854 & 3.413\% & 53s & 24.7990 & 7.263\% & 6.4m\\
\multicolumn{2}{l|}{ICAM single trajec.} & 7.8328 & 0.897\% & 2s & 10.8255& 1.139\% &\textless1s & 16.7777 & 1.551\% & 1s& 23.7976 &2.931\%  & 2s \\
\multicolumn{2}{l|}{ICAM} & 7.7991 & 0.462\% & 5s & 10.7753& 0.669\% &\textless1s & 16.6978 &1.067\%  & 4s & 23.5608 & 1.907\% &  28s\\
\multicolumn{2}{l|}{ICAM aug$\times$8} & 7.7747 & 0.148\% & 37s & \cellcolor[HTML]{D0CECE}\textbf{10.7385}& \cellcolor[HTML]{D0CECE}\textbf{0.326\%} &3s & \cellcolor[HTML]{D0CECE}\textbf{16.6488} & \cellcolor[HTML]{D0CECE}\textbf{0.771\%} & 38s & \cellcolor[HTML]{D0CECE}\textbf{23.4854} & \cellcolor[HTML]{D0CECE}\textbf{1.581\%} & 3.8m \\

\midrule
\midrule
 \multicolumn{2}{l|}{ }& \multicolumn{3}{c|}{CVRP100}  & \multicolumn{3}{c|}{CVRP200}  & \multicolumn{3}{c|}{CVRP500} & \multicolumn{3}{c}{CVRP1000}\\
 \multicolumn{2}{l|}{Method}  & Obj. & Gap & Time & Obj. & Gap & Time & Obj. & Gap & Time  & Obj. & Gap & Time\\
\midrule
\multicolumn{2}{l|}{LKH3}   &15.6465 & 0.000\% & 12h & 20.1726 & 0.000\% & 2.1h & 37.2291 & 0.000\% & 5.5h & 37.0904 & 0.000\% & 7.1h \\
\multicolumn{2}{l|}{HGS} &15.5632 & -0.533\% & 4.5h & 19.9455  & -1.126\% & 1.4h & 36.5611 & -1.794\% & 4h & 36.2884 & -2.162\% & 5.3h \\
\midrule
\multicolumn{2}{l|}{GLOP-G (LKH3)}   & $-$ & $-$ & $-$& $-$ & $-$ & $-$ & $-$ & $-$ & $-$&39.6507  & 6.903\% &1.7m\\
\midrule
\multicolumn{2}{l|}{BQ greedy}  & 16.0730 & 2.726\% & 1.8m &20.7722 & 2.972\% &10s & 38.4383 & 3.248\% & 47s & 39.2757 & 5.892\% &1.9m\\
\multicolumn{2}{l|}{LEHD greedy}    & 16.2173 & 3.648\% & 30s &20.8407 & 3.312\% & 2s& 38.4125 & 3.178\% & 17s &38.9122  & 4.912\% &1.6m\\
\midrule
\multicolumn{2}{l|}{POMO aug$\times$8} &\cellcolor[HTML]{D0CECE}\textbf{15.7544}  & \cellcolor[HTML]{D0CECE}\textbf{0.689\%} & 1.2m & 21.1542& 4.866\% &6s &44.6379  & 19.901\% & 1.2m & 84.8978 & 128.894\% &9.8m\\
\multicolumn{2}{l|}{ELG aug$\times$8}  &15.8382 & 1.225\% &4.4m  &20.6787 & 2.509\% &19s & 39.2602  & 5.456\% & 3m &41.3046 &11.362\% & 19.4m\\

\multicolumn{2}{l|}{ICAM single trajec.} & 16.1868 & 3.453\% & 2s &20.7509 & 2.867\% & \textless 1s & 37.9594 & 1.962\% & 1s &38.9709  & 5.070\% &2s\\
\multicolumn{2}{l|}{ICAM} & 15.9386 & 1.867\% & 7s & 20.5185& 1.715\% &1s & 37.6040 & 1.007\% & 5s & 38.4170 & 3.577\% &35s\\
\multicolumn{2}{l|}{ICAM aug$\times$8} & 15.8720 &1.442\% & 47s &\cellcolor[HTML]{D0CECE}\textbf{20.4334} & \cellcolor[HTML]{D0CECE}\textbf{1.293\%} &4s & \cellcolor[HTML]{D0CECE}\textbf{37.4858} & \cellcolor[HTML]{D0CECE}\textbf{0.689\%} & 42s & \cellcolor[HTML]{D0CECE}\textbf{38.2370} & \cellcolor[HTML]{D0CECE}\textbf{3.091\%} &4.5m\\

\midrule
\midrule
 \multicolumn{2}{l|}{ }& \multicolumn{3}{c|}{ATSP100}  & \multicolumn{3}{c|}{ATSP200}  & \multicolumn{3}{c|}{ATSP500} & \multicolumn{3}{c}{ATSP1000}\\
 \multicolumn{2}{l|}{Method}  & Obj. & Gap & Time & Obj. & Gap & Time & Obj. & Gap & Time  & Obj. & Gap & Time\\
\midrule
\multicolumn{2}{l|}{LKH3}   &1.5777 & 0.000\% & 17.4m &  1.6000& 0.000\% & 28s & 1.6108 & 0.000\% & 2.3m & 1.6157 & 0.000\% & 9m\\
\multicolumn{2}{l|}{OR-Tools}   & 1.8297& 15.973\% & 1.0h & 1.9209 & 20.056\% & 4m & 2.0040 & 24.410\% & 35.9m & 2.0419 & 26.379\% & 3.1h\\
\midrule
\multicolumn{2}{l|}{GLOP}   &1.7705 & 12.220\% & 23m & 1.9915 & 24.472\% & 19s & 2.207 & 36.986\% & 24s & 2.3263 & 43.980\% & 52s\\
\midrule
\multicolumn{2}{l|}{MatNet $\times$128}   & \cellcolor[HTML]{D0CECE}\textbf{1.5838}& \cellcolor[HTML]{D0CECE}\textbf{0.385\%} & 1.1h & 3.6894 & 130.588\% &4.3 m  & $-$ & $-$ &$-$  & $-$ & $-$ & $-$\\

\multicolumn{2}{l|}{ICAM}   &1.6531 & 4.782\% & 7s & \cellcolor[HTML]{D0CECE}\textbf{1.6886} & \cellcolor[HTML]{D0CECE}\textbf{5.537\%} & 1s &\cellcolor[HTML]{D0CECE}\textbf{1.7343 } & \cellcolor[HTML]{D0CECE}\textbf{7.664\%} & 5s & \cellcolor[HTML]{D0CECE}\textbf{1.8580} & \cellcolor[HTML]{D0CECE}\textbf{14.994\% }& 34s\\

\midrule
\midrule
    \multicolumn{2}{l|}{ } & \multicolumn{3}{c|}{CVRPTW100} & \multicolumn{3}{c|}{CVRPTW200} & \multicolumn{3}{c|}{CVRPTW500} & \multicolumn{3}{c}{CVRPTW1000} \\
       \multicolumn{2}{l|}{Method}    & Obj.  & Gap   & Time  & Obj.  & Gap   & Time  & Obj.  & Gap   & Time  & Obj.  & Gap   & Time \\
    \midrule
    \multicolumn{2}{l|}{HGS}   & 24.4822 & 0.000\% & 6.2h  & 41.9754 & 0.000\% & 9.8m  & 91.9256 & 0.000\% & 33.3m & 171.2468 & 0.000\% & 1.6h \\
    \midrule
    \multicolumn{2}{l|}{MTPOMO aug$\times$8} & 25.7493 & 5.176\% & 1.7m  & 46.6217 & 11.069\% & 9.3s  & 117.2123 & 27.508\% & 2.1m  & 238.2768 & 39.142\% & 16.7m \\
    \multicolumn{2}{l|}{MVMoE/4E aug$\times$8} & 25.6483 & 4.763\% & 2.3m  & 46.6037 & 11.026\% & 10.8s & 118.3341 & 28.728\% & 2.2m  & 255.4474 & 49.169\% & 17.0m \\
    \multicolumn{2}{l|}{MVMoE/4E-L aug$\times$8} & 25.6630 & 4.823\% & 2.1m  & 46.4049 & 10.552\% & 10.5s & 118.4716 & 28.878\% & 2.2m  & 255.5903 & 49.253\% & 17.5m \\
    \multicolumn{2}{l|}{ReLD-MTL aug$\times$8} & 25.5628 & 4.414\% & 2.2m  & 45.0763 & 7.387\% & 10.7s & 104.6208 & 13.810\% & 2.1m  & 203.1985 & 18.658\% & 16.1m \\
    \multicolumn{2}{l|}{ReLD-MoE aug$\times$8} & \cellcolor[HTML]{D0CECE}\textbf{25.5450} & \cellcolor[HTML]{D0CECE}\textbf{4.341\%} & 2.3m  & 45.2043 & 7.692\% & 10.9s & 105.6447 & 14.924\% & 2.1m  & 205.6633 & 20.098\% & 16.1m \\
    \midrule
    \multicolumn{2}{l|}{ICAM}  & 26.0656 & 6.468\% & 11s   & 44.7367 & 6.578\% & 2s    & 97.1875 & 5.724\% & 10s   & 179.5247 & 4.834\% & 1.1m \\
    \multicolumn{2}{l|}{ICAM aug$\times$8} & 25.8480 & 5.579\% & 1.4m  & \cellcolor[HTML]{D0CECE}\textbf{44.4468} & \cellcolor[HTML]{D0CECE}\textbf{5.888\%} & 6s    & \cellcolor[HTML]{D0CECE}\textbf{96.8040} & \cellcolor[HTML]{D0CECE}\textbf{5.307\%} & 1.2m  & \cellcolor[HTML]{D0CECE}\textbf{179.0561} & \cellcolor[HTML]{D0CECE}\textbf{4.560\%} & 8.5m \\

\bottomrule[0.5mm]
\end{tabular}
}
\label{table:uniform}
\vskip -0.1in
\end{table*}

\begin{table*}[h!]
\centering
\caption{Experimental results on cross-distribution generalization.}
\resizebox{0.98\textwidth}{!}{
\begin{threeparttable}
\begin{tabular}{ll |  cc | cc |  cc | cc }
\toprule[0.5mm]
\multicolumn{2}{l|}{ }& \multicolumn{2}{c|}{TSP1000, Rotation} & \multicolumn{2}{c|}{TSP1000, Explosion}& \multicolumn{2}{c|}{CVRP1000, Rotation} & \multicolumn{2}{c}{CVRP1000, Explosion}\\ 
\multicolumn{2}{l|}{Method}  & Obj. (Gap) & Time  & Obj. (Gap) & Time & Obj. (Gap) & Time  & Obj. (Gap) & Time\\
\midrule
\multicolumn{2}{l|}{Optimal} &  17.20 (0.00\%) & $-$ & 15.63 (0.00\%) &   $-$   &32.49 (0.00\%) & $-$& 32.31 (0.00\%) & $-$\\
\midrule
\multicolumn{2}{l|}{POMO aug$\times$8}  & 24.58 (42.84\%) & 8.5m &  22.70(45.24\%) &  8.5m  &64.22 (97.64\%) & 10.2m &59.52 (84.24\%) & 11.0m \\
\multicolumn{2}{l|}{Omni\_VRP+FS*}  & 19.53(14.30\%) & 49.9m & 17.75(13.38\%) & 49.9m   & 35.60 (10.26\%) & 56.8m & 35.25 (10.45\%)&56.8m \\
\multicolumn{2}{l|}{ELG aug$\times$8} & 19.09(10.97\%) & 15.6m &  17.37 (11.16\%) &  13.7m & 37.04(14.00\%) &20.1m  & 36.48(12.92\%) & 20.5m\\
\multicolumn{2}{l|}{ICAM}&18.97 (10.28\%) & 28s &17.35 (10.99\%) &  28s &34.72 (6.86\%) &36s &34.67 (7.31\%) & 36s\\
\multicolumn{2}{l|}{ICAM aug$\times$8} & \cellcolor[HTML]{D0CECE}\textbf{18.81(9.34\%)} & 3.8m &\cellcolor[HTML]{D0CECE}\textbf{17.17 (9.86\%) }& 3.8m   &\cellcolor[HTML]{D0CECE}\textbf{34.54 (6.28\%)} & 4.6m & \cellcolor[HTML]{D0CECE}\textbf{34.50 (6.79\%)} & 4.5m\\

\bottomrule[0.5mm]
\end{tabular}
\begin{tablenotes}
\item[\textdagger] All datasets are obtained from Omni\_VRP\cite{zhou2023omni} and contain 128 instances, and the runtime marked with an asterisk (*) is proportionally adjusted (128/1000) to match the size of our test datasets.
\end{tablenotes}
\end{threeparttable}
}
\label{table:cross_distribution}
\vskip -0.2in
\end{table*}

\paragraph{Baseline}

We compare ICAM with the following methods: 
\begin{enumerate}
    \item \textbf{Classical Solver}: Concorde~\cite{applegate2006concorde}, LKH3~\cite{LKH3}, HGS~\cite{HGS} and OR-Tools~\cite{ortools}. These methods provide nearly optimal solutions for all problem instances, allowing us to establish optimality gaps for all learning-based methods.
    \item \textbf{Constructive Neural Solvers}: POMO\cite{kwon2020pomo}, MatNet\cite{kwon2021matnet}, ELG\cite{gao2023elg}, Pointerformer~\cite{jin2023pointerformer}, Omni\_VRP~\cite{zhou2023omni},  BQ~\cite{drakulic2023bq}, LEHD~\cite{luo2023lehd}, MVMoE~\cite{zhou2024mvmoe}, ReLD~\cite{huang2025reld}, and MTPOMO~\cite{liu2024mtpomo}. As ICAM is a constructive neural solver, these baselines represent the state of the art in this category, which we aim to improve upon, particularly in terms of large-scale generalization. Notably, POMO and MatNet serve as the backbones for our ICAM models, making it crucial to include them in the comparison despite their poor performance.
    \item \textbf{Two-stage Neural Solvers}: Att-GCN+MCTS~\cite{fu2021attgcn_mcts}, DIMES~\cite{qiu2022dimes}, SO~\cite{cheng2023select}, DIFUSCO~\cite{sun2023difusco}, H-TSP~\cite{pan2023htsp}, T2T~\cite{li2024T2T} and GLOP~\cite{ye2023glop}. This group represents an alternative paradigm for solving large-scale routing problems, often by combining learning with a well-designed decomposition or search strategy. Comparing our method with these methods shows that the constructive ICAM remains highly competitive with these more complex, multi-stage solvers.
\end{enumerate}

\begin{table}[t]
\centering
\caption{Comparison on TSPLIB~\cite{reinelt1991tsplib} with scale $\leq 5,000$. }
\resizebox{0.99\columnwidth}{!}{
\begin{threeparttable}
\begin{tabular}{ll | c| c| c| c |  c }
\toprule[0.5mm]
\multicolumn{2}{l|}{ }& $ N \leq1000$ & $1000 $ \textless $ N \leq2000$ & $2000 $ \textless $ N \leq 3000$ &  $3000 $ \textless $ N \leq 5000$ & \\
 \multicolumn{2}{l|}{Method }& (48 instances) & (15 instances) &(4 instances) &(3 instances)&Avg.time \\

\midrule
\multicolumn{2}{l|}{LEHD greedy}&\cellcolor[HTML]{D0CECE}\textbf{2.51\%} &10.54\%  & 10.93\% &15.35\% & 4.4s   \\
\multicolumn{2}{l|}{BQ greedy} &3.04\%&\cellcolor[HTML]{D0CECE}\textbf{9.72\% } & 11.58\% &22.88\%& 4.3s   \\
\midrule
\multicolumn{2}{l|}{POMO aug$\times$8} &9.31\%&60.27\%  & 61.28\% &84.90\%& 15.9s   \\
\multicolumn{2}{l|}{ELG aug$\times$8} &2.98\%&11.79\%  & 9.05\% & OOM\textdagger & $-$   \\
\multicolumn{2}{l|}{ICAM} &5.95\%&13.28\%  & 9.88\% & 14.95\%&1.0s  \\
\multicolumn{2}{l|}{ICAM aug$\times$8} &4.00\%&11.52\%  & \cellcolor[HTML]{D0CECE}\textbf{8.86\%} &\cellcolor[HTML]{D0CECE}\textbf{14.01\%}& 6.5s   \\
\bottomrule[0.5mm]
\end{tabular}
\begin{tablenotes}
\large{
\item[\textdagger] Instance fnl4461 cannot be solved due to the OOM issue.
}

\end{tablenotes}
\end{threeparttable}
}
\label{table:TSPLIB_all}
\vskip -0.2in
\end{table}
\paragraph{Metrics and Inference}
We use objective values, optimality gaps, and total inference times to evaluate each method. Specifically, the optimality gap measures the discrepancy between the solutions generated by learning and non-learning methods and the optimal solutions, which are obtained using LKH3 for all problems. Note that times for classical solvers, which run on a single CPU, and for learning-based methods, which utilize GPUs, are inherently different. Therefore, these times should not be directly compared.

For most neural solvers, we directly execute the source code provided by the authors using default settings. Note that the results marked with an asterisk (*) are directly obtained from the corresponding papers. For TSPs and CVRPs, following \cite{kwon2020pomo}, we report three types of results: using a single trajectory, the best result from multiple trajectories, and results derived from instance augmentation. For ATSPs, we remove instance augmentation and report only the best result from multiple trajectories using a greedy strategy, rather than the sampled ones adopted by MatNet.

\subsection{Performance Evaluation}
\label{sec:performance_evaluation}
\paragraph{Results on VRPs with Scale $\leq 1,000$}
The experimental results with uniform distribution and scale $\leq 1,000$ are reported in \cref{table:uniform}. Our method stands out for consistently delivering superior inference performance, complemented by remarkably fast inference times, across various problem instances. Although it cannot surpass Att-GCN+MCTS on TSP100, POMO on CVRP100, MatNet on ATSP100, and ReLD on CVRPTW100, the time it consumes is significantly less, such as MatNet requires over an hour compared to our $7$s. On TSP1000 instances, our model impressively reduces the optimality gap to less than 3\% in just $2$ seconds. When switching to a multi-greedy strategy, the optimality gap further narrows to 1.9\% in $30$ seconds. With the instance augmentation, ICAM can achieve the optimality gap of 1.58\% in less than $4$ minutes. To the best of our knowledge, for instances with up to $1,000$ nodes, ICAM shows state-of-the-art generalization performance among all RL-based constructive methods in the four different tested problems. The comprehensive results demonstrate that ICAM can consistently generate highly efficient solutions for various route planning scenarios in real-time, which can directly translate to improved service efficiency and reduced energy consumption in modern ITS.

\begin{table}[t]
\centering
\caption{Comparison on CVRPLIB Set-X\cite{uchoa2017new}.}
\resizebox{0.98\columnwidth}{!}{
\begin{tabular}{ll | c| c |  c | c | c }
\toprule[0.5mm]
 \multicolumn{2}{l|}{ }& $N \leq200$ &$200 \textless N \leq 500$ & $500 \textless N \leq1000$& Total& \\
 \multicolumn{2}{l|}{Method }& (22 instances) &(46 instances) &(32 instances)&(100 instances)& Avg.time\\
\midrule
\multicolumn{2}{l|}{LEHD greedy}   & 11.35\% &9.45\% &17.74\% &12.52\% & 1.58s \\
\multicolumn{2}{l|}{BQ greedy*}   & $-$ &$-$ &$-$ &9.94\% & $-$ \\
\midrule
\multicolumn{2}{l|}{POMO aug$\times$8}   & 6.90\% & 15.04\% &40.81\% &21.49\% & 1.0s \\ 
\multicolumn{2}{l|}{ELG aug$\times$8}   & 4.51\% &5.52\% &7.80\% &6.03\% & 2.6s \\
\multicolumn{2}{l|}{ICAM}   & 5.14\% &4.44\% &5.17\% &4.83\% & 0.4s \\ 
\multicolumn{2}{l|}{ICAM aug$\times$8}   & \cellcolor[HTML]{D0CECE}\textbf{4.41\%} &\cellcolor[HTML]{D0CECE}\textbf{3.92\% }&\cellcolor[HTML]{D0CECE}\textbf{4.70\%}&\cellcolor[HTML]{D0CECE}\textbf{4.28\%} & 0.6s \\  

\bottomrule[0.5mm]
\end{tabular}
}
\label{table:CVRPLIB_set_x}
\vskip -0.1in
\end{table}

\begin{table}[t]
\centering
\caption{Comparison on CVRPLIB Set-XXL~\cite{arnold2019cvrplib_xxl} with scale $\in [3000,7000]$. }
\resizebox{0.99\columnwidth}{!}{
\begin{tabular}{ll | c| c |  c | c  | c | c }
\toprule[0.5mm]
 \multicolumn{2}{l|}{ }& Leuven1 & Leuven2 & Antwerp1 & Antwerp2 & Total & \\
 \multicolumn{2}{l|}{Method }&($N=3000$)&($N=4000$) & ($N=6000$) &($N=7000$) &$N \in [3000,7000]$&Avg.time\\
\midrule
\multicolumn{2}{l|}{LEHD greedy}   &16.60\% &34.86\% & 14.66\% &22.77\% & 22.22\%& 155.3s\\
\multicolumn{2}{l|}{BQ greedy}   &18.53\% & 30.70\% & 16.48\% &27.67\%  & 23.34\%& 30.0s\\
\midrule
\multicolumn{2}{l|}{POMO aug$\times$8}   &460.32\% & 202.17\%  & OOM &OOM & $-$& $-$\\ 

\multicolumn{2}{l|}{ELG aug$\times$8*}   &10.77\% &21.80\% & 10.70\% &17.69\% & 15.24\%& $-$\\ 
\multicolumn{2}{l|}{ICAM}   &9.22\%&15.09\%  & 8.00\% &21.66\% & 13.49\%& 39.9s\\ 
\multicolumn{2}{l|}{ICAM aug$\times$8}   &\cellcolor[HTML]{D0CECE}\textbf{9.22\% }&\cellcolor[HTML]{D0CECE}\textbf{15.09\%} & \cellcolor[HTML]{D0CECE}\textbf{7.45\%} &\cellcolor[HTML]{D0CECE}\textbf{14.31\%} & \cellcolor[HTML]{D0CECE}\textbf{11.52\%}&  253.0s\\ %
\bottomrule[0.5mm]
\end{tabular}
}
\label{table:CVRPLIB_set_xxl}
\vskip -0.2in
\end{table}
\paragraph{Results on Cross-distribution VRP Instances} 
We use the TSP/CVRP$1,000$ datasets with rotation and explosion distributions to evaluate the cross-distribution performance of ICAM. As shown in \cref{table:cross_distribution}, ICAM can still achieve the best performance on specific distribution instances and the fastest speed of all comparable models. These results confirm that the same adaptation function $f(N,d_{ij})$ can perform well across problem instances with different distributions. Our experiments on two non-uniform distributions serve as a critical case study for diverse city topologies, verifying ICAM's robustness in adapting to varying geographical layouts, ensuring reliable deployment across diverse transportation networks.

\paragraph{Results on Benchmark Dataset} 
We further evaluate the performance using well-known benchmark datasets from TSPLIB~\cite{reinelt1991tsplib} with scale $\leq 5000$ (see \cref{table:TSPLIB_all}), CVRPLIB Set-X~\cite{uchoa2017new} with scale $\leq 1000$ (see \cref{table:CVRPLIB_set_x}), and Set-XXL~\cite{arnold2019cvrplib_xxl} with scale $\in [3000,7000]$ (see \cref{table:CVRPLIB_set_xxl}). In TSPLIB datasets, although ICAM exhibits marginally inferior performance compared to SL-based models (i.e., BQ and LEHD) with scale $\leq 2000$,  it demonstrates superior solution quality in larger-scale instances exceeding $2000$ nodes, which better reflect real-world scenarios. In Set-X and Set-XXL benchmarks, ICAM consistently outperforms all baseline models across instances with scale $\in [100,7000]$. These results further highlight the robust generalization capability of ICAM.

\paragraph{Applicability to Real-World Systems}
\label{sec:real_world}
Following the experimental protocols in recent ITS studies~\cite{li2025map1,tang2025map3,jia2025arc_tits}, we conduct a comprehensive case study using real-world road networks derived from the OpenStreetMap (OSM) platform~\cite{OpenStreetMap2025}. Specifically, we employ the dataset provided by \cite{son2025rrnco} and select eight representative cities (Beijing, Chicago, London, Melbourne, Seoul, Sydney, Tokyo, Toronto) across four continents (Asia, Europe, North America, and Oceania) to ensure diverse urban layout structures. 

\begin{table*}[t]
  \centering
  \caption{Zero-shot generalization on real-world CVRP instances derived from OSM. The notation 'City-$n$' indicates an instance with $n$ customer nodes sampled from the original map. All methods employ $\times 8$ instance augmentation. }
  \resizebox{0.98\textwidth}{!}{
    \begin{tabular}{l|cc|cc|cc|cc}
    \toprule[0.5mm]
    \multirow{2}[1]{*}{Method} & \multicolumn{2}{c|}{Beijing-100 } & \multicolumn{2}{c|}{Beijing-500} & \multicolumn{2}{c|}{Chicago-100} & \multicolumn{2}{c}{Chicago-500} \\
          & Gap   & Avg.Time & Gap   & Avg.Time & Gap   & Avg.Time & Gap   & Avg.Time \\
    \midrule
    HGS   & 0.000\% & 10m   & 0.000\% & 10m   & 0.000\% & 10m   & 0.000\% & 10m \\
    \midrule
    POMO  & 20.594\% & 0.011s & 73.911\% & 0.58s & 16.597\% & 0.011s & 70.047\% & 0.58s \\
    Omni\_VRP & 21.443\% & 0.010s & 50.124\% & 0.56s & 17.793\% & 0.010s & 45.770\% & 0.56s \\
    ELG   & 11.801\% & 0.03s & 15.618\% & 1.41s & 10.252\% & 0.03s & 15.000\% & 1.42s \\
    ICAM  & \cellcolor[HTML]{D0CECE}\textbf{7.880\%} & \cellcolor[HTML]{D0CECE}\textbf{0.009s} & \cellcolor[HTML]{D0CECE}\textbf{9.044\%} & \cellcolor[HTML]{D0CECE}\textbf{0.28s} & \cellcolor[HTML]{D0CECE}\textbf{7.107\%} & \cellcolor[HTML]{D0CECE}\textbf{0.009s} & \cellcolor[HTML]{D0CECE}\textbf{8.530\%} & \cellcolor[HTML]{D0CECE}\textbf{0.28s} \\
    \midrule
    \midrule
    \multirow{2}[1]{*}{Method} & \multicolumn{2}{c|}{London-100 } & \multicolumn{2}{c|}{London-500} & \multicolumn{2}{c|}{Melbourne-100} & \multicolumn{2}{c}{Melbourne-500} \\
          & Gap   & Avg.Time & Gap   & Avg.Time & Gap   & Avg.Time & Gap   & Avg.Time \\
    \midrule
    HGS   & 0.000\% & 10m   & 0.000\% & 10m   & 0.000\% & 10m   & 0.000\% & 10m \\
    \midrule
    POMO  & 16.080\% & 0.011s & 64.053\% & 0.58s & 13.410\% & 0.011s & 54.155\% & 0.58s \\
    Omni\_VRP & 17.589\% & 0.010s & 35.243\% & 0.56s & 14.777\% & 0.010s & 37.308\% & 0.56s \\
    ELG   & 10.018\% & 0.03s & 14.631\% & 1.42s & 8.987\% & 0.03s & 14.899\% & 1.42s \\
    ICAM  & \cellcolor[HTML]{D0CECE}\textbf{7.092\%} & \cellcolor[HTML]{D0CECE}\textbf{0.009s} & \cellcolor[HTML]{D0CECE}\textbf{8.828\%} & \cellcolor[HTML]{D0CECE}\textbf{0.28s} & \cellcolor[HTML]{D0CECE}\textbf{6.626\%} & \cellcolor[HTML]{D0CECE}\textbf{0.009s} & \cellcolor[HTML]{D0CECE}\textbf{8.530\%} & \cellcolor[HTML]{D0CECE}\textbf{0.28s} \\
    \midrule
    \midrule
    \multirow{2}[2]{*}{Method} & \multicolumn{2}{c|}{Seoul-100 } & \multicolumn{2}{c|}{Seoul-500} & \multicolumn{2}{c|}{Sydney-100} & \multicolumn{2}{c}{Sydney-500} \\
          & Gap   & Avg.Time & Gap   & Avg.Time & Gap   & Avg.Time & Gap   & Avg.Time \\
    \midrule
    HGS   & 0.000\% & 10m   & 0.000\% & 10m   & 0.000\% & 10m   & 0.000\% & 10m \\
    \midrule
    POMO  & 13.378\% & 0.011s & 55.203\% & 0.58s & 16.562\% & 0.011s & 62.726\% & 0.58s \\
    Omni\_VRP & 15.053\% & 0.010s & 35.585\% & 0.56s & 18.129\% & 0.010s & 43.819\% & 0.56s \\
    ELG   & 8.851\% & 0.03s & 13.450\% & 1.42s & 10.483\% & 0.03s & 15.707\% & 1.42s \\
    ICAM  & \cellcolor[HTML]{D0CECE}\textbf{7.347\%} & \cellcolor[HTML]{D0CECE}\textbf{0.009s} & \cellcolor[HTML]{D0CECE}\textbf{8.281\%} & \cellcolor[HTML]{D0CECE}\textbf{0.28s} & \cellcolor[HTML]{D0CECE}\textbf{7.117\%} & \cellcolor[HTML]{D0CECE}\textbf{0.009s} & \cellcolor[HTML]{D0CECE}\textbf{8.962\%} & \cellcolor[HTML]{D0CECE}\textbf{0.28s} \\
    \midrule
    \midrule
    \multirow{2}[2]{*}{Method} & \multicolumn{2}{c|}{Tokyo-100 } & \multicolumn{2}{c|}{Tokyo-500} & \multicolumn{2}{c|}{Toronto-100} & \multicolumn{2}{c}{Toronto-500} \\
          & Gap   & Avg.Time & Gap   & Avg.Time & Gap   & Avg.Time & Gap   & Avg.Time \\
    \midrule
    HGS   & 0.000\% & 10m   & 0.000\% & 10m   & 0.000\% & 10m   & 0.000\% & 10m \\
    \midrule
    POMO  & 12.313\% & 0.011s & 51.722\% & 0.58s & 8.146\% & 0.011s & 42.682\% & 0.58s \\
    Omni\_VRP & 13.615\% & 0.010s & 35.689\% & 0.56s & 9.650\% & 0.010s & 24.194\% & 0.56s \\
    ELG   & 8.474\% & 0.03s & 13.704\% & 1.42s & 5.918\% & 0.03s & 11.089\% & 1.42s \\
    ICAM  & \cellcolor[HTML]{D0CECE}\textbf{7.042\%} & \cellcolor[HTML]{D0CECE}\textbf{0.009s} & \cellcolor[HTML]{D0CECE}\textbf{9.099\%} & \cellcolor[HTML]{D0CECE}\textbf{0.28s} & \cellcolor[HTML]{D0CECE}\textbf{5.441\%} & \cellcolor[HTML]{D0CECE}\textbf{0.009s} & \cellcolor[HTML]{D0CECE}\textbf{6.500\%} & \cellcolor[HTML]{D0CECE}\textbf{0.28s} \\
    \bottomrule[0.5mm]
    \end{tabular}%
    }
  \label{tab:real_world_results}%
\end{table*}%
\begin{figure}[h!]
    \centering
    \begin{subfigure}{0.24\textwidth}
        \centering
        \includegraphics[width=\textwidth]{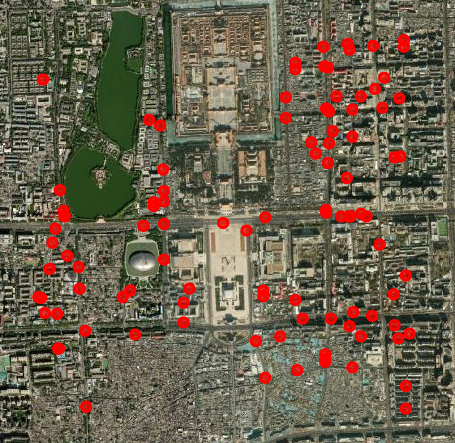} 
        \caption{Beijing-100}
    \end{subfigure}
    \hfill
    \begin{subfigure}{0.24\textwidth}
        \centering
        \includegraphics[width=\textwidth]{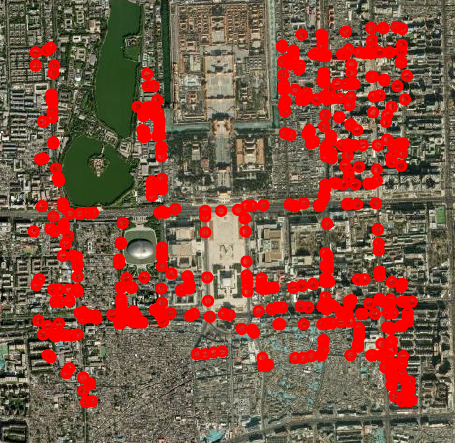} 
        \caption{Beijing-500}
    \end{subfigure}
    \caption{Visualization of instances sampled from different scales in Beijing. For more visualizations of different cities, please see Appendix \ref{append:real_world}.}
    \label{fig:real_world_visual}
    \vspace{-20pt}
\end{figure}
Following the methodology outlined in \cite{son2025rrnco}, we have randomly sampled subgraphs to generate datasets with 100 and 500 customer locations (128 instances per dataset). To better reflect practical logistics conditions, we focus on the more realistic CVRP. Node demands and vehicle capacities are set according to the standard configurations described in \cite{luo2023lehd}. As illustrated in Figure \ref{fig:real_world_visual}, the acquired real-world data exhibits typical distribution characteristics that differ significantly from synthetic uniform data, providing a meaningful testbed for real-world applicability. 

In this experiment, we have compared ICAM with mainstream RL-based constructive solvers. All models are trained exclusively on synthetic uniformly distributed instances and are not fine-tuned on any real-world problem. Near-optimal solutions obtained by the HGS heuristic~\cite{HGS} are served as references for calculating optimality gaps. The results presented in Table \ref{tab:real_world_results} demonstrate that ICAM can better address the distribution shift from synthetic Euclidean spaces to real-world road network distances without any retraining, and significantly outperform other learning-based methods. Notably, ICAM delivers high-quality solutions with extremely fast inference speeds, making it a highly scalable solution for real-time intelligent transportation systems. These results demonstrate ICAM's strong adaptability and its potential for deployment in practical ITS.

\section{Discussions}
\label{sec:discussions}
\subsection{AAFM vs. MHA}
\label{sec:aft_mha}
In language modeling, the relation (e.g., semantic difference) between two tokens is difficult to represent directly by position bias $w_{i,j}$. However, in routing problems, the relation between two nodes can be directly represented by only the distance information computed from the node coordinates, just as a traditional heuristic solver (e.g., LKH3~\cite{LKH3}) can solve a specific instance by only inputting the distance-based adjacency matrix. In classic neural vehicle routing solvers using MHA, e.g., POMO\cite{kwon2020pomo}, the relation between two nodes is computed by mapping the node coordinates into a high-dimensional hidden space. In short, MHA cannot directly take advantage of the pair-wise distances between nodes. However, explicit relative position information is valuable for achieving better performance for routing problems.

To investigate the effectiveness of AAFM compared to MHA in information integration, we train a new ICAM that replaces AAFM with the standard MHA, denoted as ICAM-MHA. ICAM-MHA is trained in exactly the same settings, including three-stage training, the adaptation function, and hyperparameters. The only difference between the two models is the attention mechanism. Considering the special design of MHA, the way that we integrate the adaptation bias matrix $A$ with Self-Attention in MHA can be expressed as
\begin{equation}
      \mathrm{Attention}^{*}(Q, K, V) = \mathrm{softmax}\left(\frac{QK^{\mathrm{T}}}{\sqrt{d_k}} + A\right)V,
    \label{eq:mha_adaptation}
\end{equation}

\begin{table}[h!]
\centering
\caption{Comparison of the AAFM and MHA on TSP instances with different scales.}
\resizebox{0.49\textwidth}{!}{
\begin{tabular}{ll | ccc | ccc }
\toprule[0.5mm]
\multicolumn{2}{l|}{ }& \multicolumn{3}{c|}{TSP100 }  & \multicolumn{3}{c}{TSP1000}\\ 
\multicolumn{2}{l|}{Method}  & Obj. & Gap & Time & Obj. & Gap & Time \\
\midrule
\multicolumn{2}{l|}{Concorde} &7.7632 & 0.000\% & 34m & 23.1199 & 0.000\% & 7.8h \\

\midrule
\multicolumn{2}{l|}{ICAM-MHA} &7.8061 & 0.552\% & 10s & 23.7193& 2.593\% &1.5m\\

\multicolumn{2}{l|}{ICAM} & \cellcolor[HTML]{D0CECE}\textbf{7.7991 }& \cellcolor[HTML]{D0CECE}\textbf{0.462\%} & \cellcolor[HTML]{D0CECE}\textbf{5s} & \cellcolor[HTML]{D0CECE}\textbf{23.5608} & \cellcolor[HTML]{D0CECE}\textbf{1.907\% }&  \cellcolor[HTML]{D0CECE}\textbf{28s}\\
\bottomrule[0.5mm]
\end{tabular}
}
\label{table:compare_AFT_MHA}
\vskip -0.05in
\end{table}

As shown in \cref{table:compare_AFT_MHA}, ICAM-MHA also has good large-scale generalization performance. This result again demonstrates the effectiveness of the proposed adaptation function and three-stage training scheme. In addition, we can observe that replacing MHA with AAFM can further improve performance while significantly reducing running time. The advantages of ICAM over ICAM-MHA become more significant as the problem scale increases. The good scalability performance of ICAM may stem from the ability of AAFM to integrate instance-conditioned information more efficiently.

\begin{figure*}[h!]
    \centering
    \begin{subfigure}{0.24\textwidth}
        \centering
        \includegraphics[width=\textwidth]{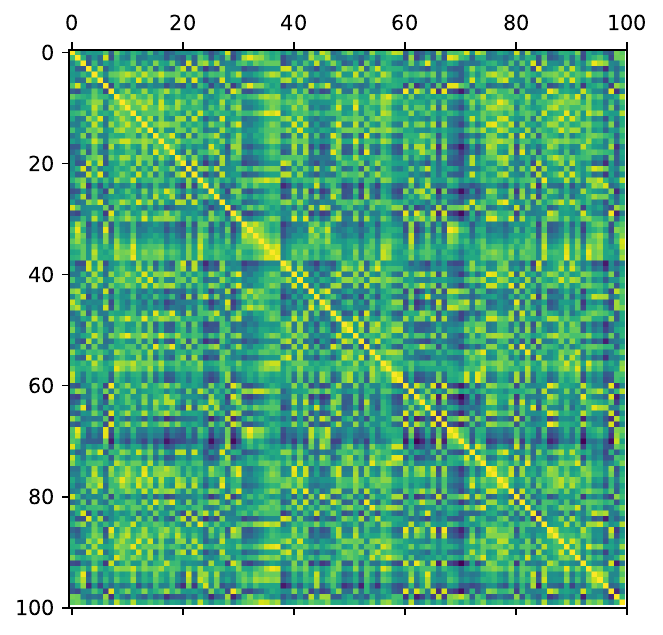} 
        \caption{\centering Pair-wise distance \newline (TSP100)}
        \label{subfig:dist_hm_tsp100}
    \end{subfigure}
    \hfill
    \begin{subfigure}{0.24\textwidth}
        \centering
        \includegraphics[width=\textwidth]{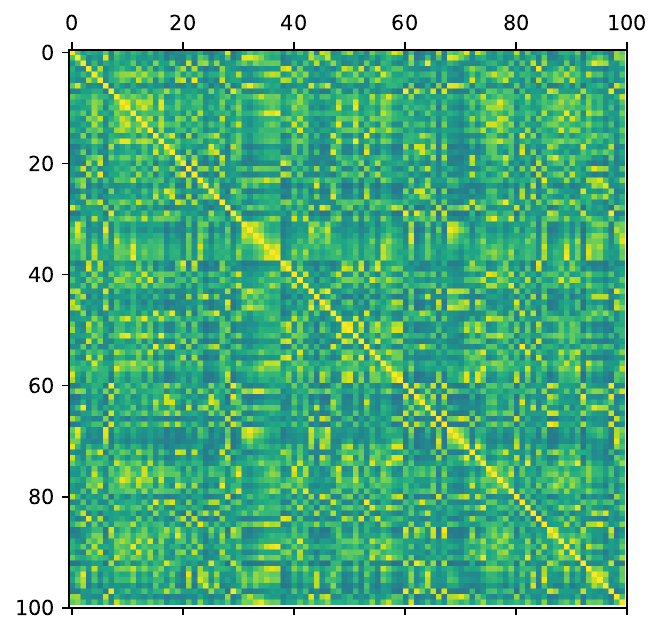}
        \caption{\centering Pair-wise similarity\newline (ICAM, TSP100)}
        \label{subfig:cosine_icam_tsp100}
    \end{subfigure}
    \hfill
     \begin{subfigure}{0.24\textwidth}
        \centering
        \includegraphics[width=\textwidth]{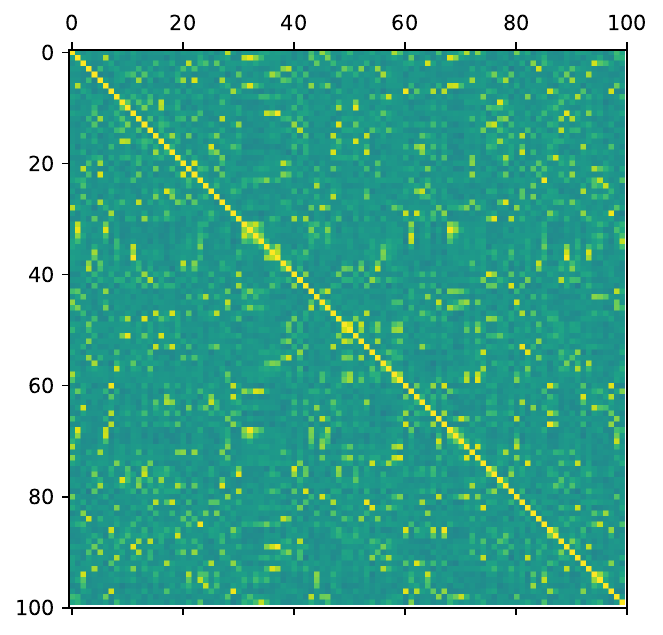}
        \caption{\centering Pair-wise similarity\newline (ELG, TSP100)}
        \label{subfig:cosine_elg_tsp100}
    \end{subfigure}
    \hfill
     \begin{subfigure}{0.24\textwidth}
        \centering
        \includegraphics[width=\textwidth]{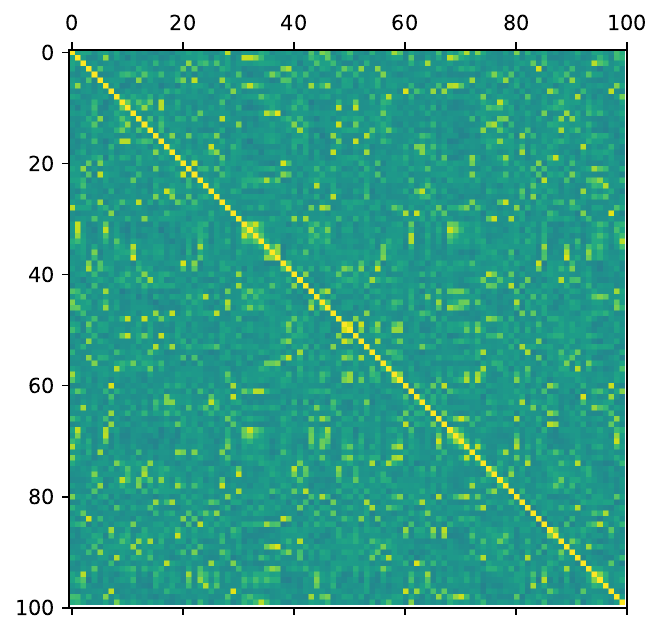}
        \caption{\centering Pair-wise similarity\newline (POMO,TSP100)}
        \label{subfig:cosine_pomo_tsp100}
    \end{subfigure}
    \hfill
    \begin{subfigure}{0.24\textwidth}
        \centering
        \includegraphics[width=\textwidth]{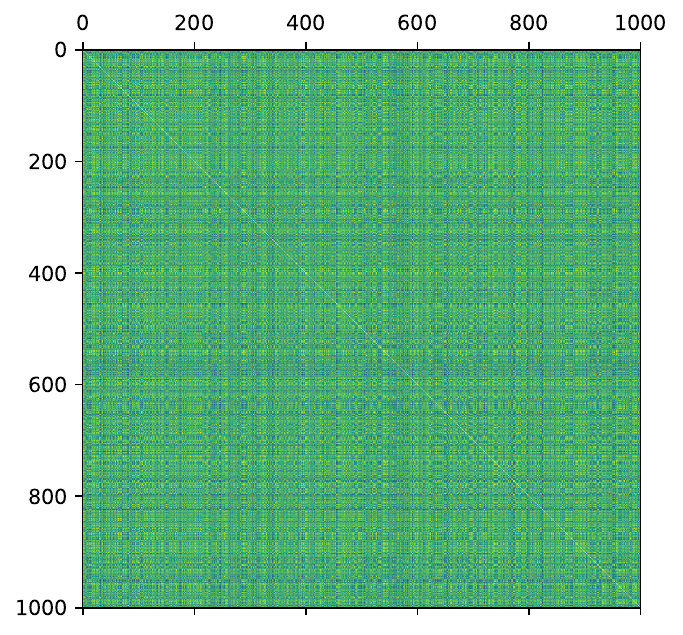} 
        \caption{\centering Pair-wise distance \newline (TSP1000)}
        \label{subfig:dist_hm_tsp1000}
    \end{subfigure}
    \hfill
    \begin{subfigure}{0.24\textwidth}
        \centering
        \includegraphics[width=\textwidth]{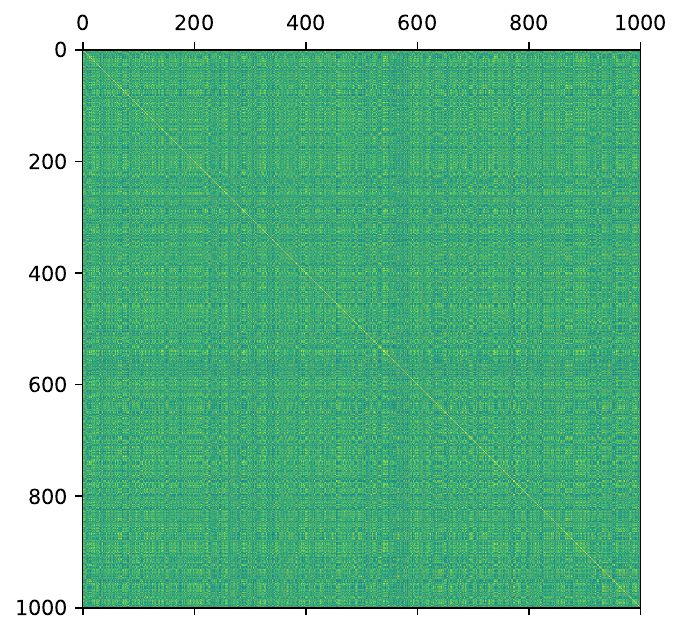}
        \caption{\centering Pair-wise similarity\newline (ICAM, TSP1000)}
        \label{subfig:cosine_icam_tsp1000}
    \end{subfigure}
    \hfill
     \begin{subfigure}{0.24\textwidth}
        \centering
        \includegraphics[width=\textwidth]{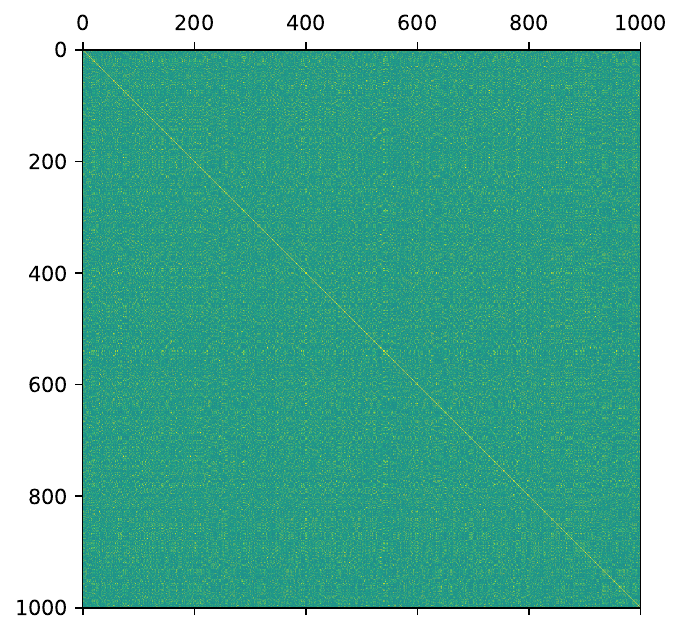}
        \caption{\centering Pair-wise similarity\newline (ELG, TSP1000)}
        \label{subfig:cosine_elg_tsp1000}
    \end{subfigure}
    \hfill
     \begin{subfigure}{0.24\textwidth}
        \centering
        \includegraphics[width=\textwidth]{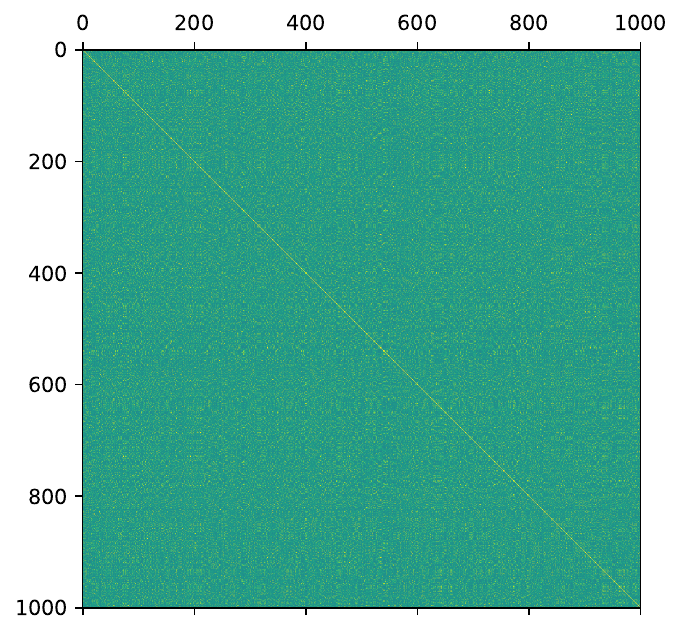}
        \caption{\centering Pair-wise similarity\newline (POMO,TSP1000)}
        \label{subfig:cosine_pomo_tsp1000}
    \end{subfigure}

    \caption{Comparison of cosine similarity between node embeddings generated by the encoders of different models and actual pair-wise distance with different scales. It is noteworthy that darker shades indicate lower similarity.  For a more comprehensive comparison across various scales, please see Appendix \ref{append:capturing_features}.}
    \label{fig:tsp_hm_short}
    \vskip -0.2in
\end{figure*}

\subsection{Comparison with Learning-enhanced Metaheuristic}
\label{sec:meta_heuristic}
To further demonstrate the performance of ICAM, we compared ICAM against a representative learning-enhanced metaheuristic GANCP~\cite{sobhanan2025genetic}, using its provided default settings. Since the efficacy of GANCP relies on the consistency between training and testing distributions, we utilize the CVRPLib Set-X benchmark~\cite{uchoa2017new} for evaluation. This benchmark aligns with the instance generation rules used to train GANCP, thereby ensuring a fair comparison. We report the best objective value selected from the 5 candidate solutions generated by population-based GANCP.

\begin{table}[htbp]
\centering
\caption{Comparison between ICAM and GANCP~\cite{sobhanan2025genetic} on CVRPLIB Set-X\cite{uchoa2017new}. }
\resizebox{0.98\columnwidth}{!}{
\begin{tabular}{ll | c| c |  c | c | c }
\toprule[0.5mm]
 \multicolumn{2}{l|}{ }& $N \leq200$ &$200 \textless N \leq 500$ & $500 \textless N \leq1000$& Total& \\
 \multicolumn{2}{l|}{Method }& (22 instances) &(46 instances) &(32 instances)&(100 instances)& Avg.time\\
\midrule
\multicolumn{2}{l|}{GANCP} & 4.94\%& 5.75\%& 6.12\%& 5.69\% & 20.5s\\
\multicolumn{2}{l|}{ICAM aug$\times$8}   & \cellcolor[HTML]{D0CECE}\textbf{4.41\%} &\cellcolor[HTML]{D0CECE}\textbf{3.92\% }&\cellcolor[HTML]{D0CECE}\textbf{4.70\%}&\cellcolor[HTML]{D0CECE}\textbf{4.28\%} & \cellcolor[HTML]{D0CECE}\textbf{0.6s} \\ 

\bottomrule[0.5mm]
\end{tabular}
}
\label{table:gancp}
\vskip -0.1in
\end{table}

The results presented in Table \ref{table:gancp} demonstrate that ICAM consistently outperforms GANCP across all scale ranges, achieving a lower average overall optimality gap (4.28\% vs. 5.69\%). Notably, ICAM is orders of magnitude faster, requiring only 0.6 seconds on average, compared to 20.5 seconds for GANCP.

\subsection{Further Improvement of Solutions}
\label{sec:2-opt}

\begin{table}[h!]
\centering
\caption{Comparison between the original ICAM and the results augmented with 2-opt.}
\resizebox{0.48\textwidth}{!}{
\begin{tabular}{ll | cc | cc |  cc | cc }
\toprule[0.5mm]

\multicolumn{2}{l|}{ }& \multicolumn{2}{c|}{TSP100 }  & \multicolumn{2}{c|}{TSP200}  & \multicolumn{2}{c|}{TSP500} & \multicolumn{2}{c}{TSP1000}\\ 
\multicolumn{2}{l|}{Method}   & Gap & Time & Gap & Time  & Gap & Time   & Gap & Time\\
\midrule
\multicolumn{2}{l|}{LKH3}    & 0.000\% & 56m  & 0.000\% & 4m &  0.000\% & 32m  & 0.000\% & 8.2h \\
\midrule

\multicolumn{2}{l|}{ICAM aug$\times$8}  & 0.148\% & 37s & 0.326\% &3s  & 0.771\% & 38s& 1.581\% & 3.8m \\
\multicolumn{2}{l|}{ICAM aug$\times$8+2OPT}   & \cellcolor[HTML]{D0CECE}\textbf{0.134\%} & 47s  & \cellcolor[HTML]{D0CECE}\textbf{0.291\%} & 4s  & \cellcolor[HTML]{D0CECE}\textbf{0.695\%} & 42s  & \cellcolor[HTML]{D0CECE}\textbf{1.327\%} & 4m \\ 

\bottomrule[0.5mm]
\end{tabular}
}
\label{table:tsp_uniform_2opt}
\end{table}

To further explore the performance limits of our model, we apply a 2-opt local search to refine the solutions generated by ICAM (specifically the ICAM aug$\times$8 version), following a common practice in neural routing solvers~\cite{sun2023difusco,li2024T2T}. We set the maximum number of iterations to 1000 to ensure that the local search had sufficient opportunity to converge. 

As presented in Table \ref{table:tsp_uniform_2opt}, with the aid of 2-opt, ICAM achieves better solutions (e.g., reducing the gap from 1.581\% to 1.327\% on TSP1000). Notably, we also observe that the 2-opt search approach typically stops within $100$ iterations (i.e., it already converged), even when allowed up to $1000$ iterations, which suggests that the initial greedy-decoded solutions are already of high quality and leave limited room for simple local refinement. Even when augmented with 2‑opt, ICAM remains orders of magnitude faster than LKH3 (e.g., 4 minutes versus 8.2 hours for TSP1000). 

These results further validate the robustness of ICAM. It delivers high-quality solutions quickly and can be flexibly combined with lightweight local search to provide better solutions with a small computational overhead.

\subsection{Capturing Instance-specific Features} 
\label{sec:capturing_features}
Given the diverse variations in patterns and geometric structures across different scales, we argue that instance-conditioned adaptation is crucial for improving the generalization of RL-based neural routing solvers. While various approaches have been explored for integrating auxiliary information, current RL-based methods still struggle to achieve a satisfying generalization performance, especially for large-scale instances. The RL-based models generally adopt a heavy encoder and light decoder structure, where the quality of node embeddings generated by the encoder plays a pivotal role in overall performance. Given the diverse geometric structures and patterns of instances across different scales, we argue that the ability of node embeddings to adaptively capture instance-specific features across varying-scale instances is crucial for improving generalization performance.

To verify whether the node embeddings can effectively capture instance-specific features, we calculate the correlation between pair-wise node features using the cosine similarity between node embeddings generated by the encoder. The cosine similarity calculation is defined as:
\begin{equation}
\begin{split}
\mathrm{Similarity}(e_i, e_j) & = \dfrac{e_i \cdot e_j}{\max\left(\Vert e_i \Vert _2 \cdot \Vert e_j \Vert _2,  \epsilon\right)} \\& = \frac{\sum_{k=1}^{dim} e_{i,k} \times e_{j,k}}{\max\left(\sqrt{\sum_{k=1}^{dim} e_{i,k}^2} \times \sqrt{\sum_{k=1}^{dim} e_{j,k}^2},\epsilon\right)}
\label{eq:cosine_sim}
\end{split}
\end{equation}
where $e_i$ and $e_j$ represent the embeddings generated by the encoder of node $i$ and node $j$, respectively, $dim$ is the embedding dimension, $\epsilon$ is a small value to avoid division by zero ($\epsilon = 1e-8$ in this work). It is easy to check the range of $\mathrm{Similarity}(e_i, e_j)$ is $[-1,1]$. A similarity value $1$ means the two compared embeddings are exactly the same, a value $-1$ means they are in the opposite direction. Once we have this similarity matrix for embeddings, we can compare it with the distance matrix of nodes to check whether they share similar patterns. For an easy visualization comparison, we can calculate the inverse distance matrix with the component $\hat d_{ij} = \max_{i,j}{d_{ij}} - d_{ij}$ and further normalize the whole matrix to the range $[-1,1]$ via $\hat d_{ij} = 2\cdot \frac{\hat d_{ij}}{\max_{i,j}{\hat d_{ij}}} - 1$, where a value $\hat d_{ij} = 1$ means node $i$ and node $i$ are at exactly the same location, and $\hat d_{ij} = -1$ means they are far away from each other. In this way, if the node embeddings can successfully capture instance-specific features, their similarity matrix should exhibit some similar patterns to the normalized inverse distance matrix.

We conduct a case study on TSP to demonstrate the instance-conditioned adaptation ability of different models, as shown in \cref{fig:tsp_hm_short}. According to the results, the representative RL-based models (i.e., ELG and POMO) all fail to effectively capture instance-specific features in their node embeddings. On the other hand, our proposed ICAM can generate instance-conditioned node embeddings, of which the embedding correlation matrix shares similar patterns with the original distance matrix. These results clearly demonstrate that ICAM can successfully capture instance-specific features in its embeddings, leading to its promising generalization.

\subsection{Complexity Analysis}
As shown in \cref{table:gpu_memory_time}, we report the model size, memory usage per instance, and total inference time for different RL-based constructive models. We report the complexity of the model under adopting the multi-greedy strategy. For GPU memory, we report the average GPU memory usage per instance of each method for each problem. Due to our 12-layer encoder, we have more parameters than POMO and ELG. However, since the heavy encoder is only called once for solution construction, our ICAM method achieves the lowest memory usage and the fastest inference time for all TSPs.
\begin{table}[h!]
\centering  
\caption{Complexity analysis between ICAM and existing works. "Avg.memory" represents the average memory usage per instance. $N$ and $k$ denote the scale and the number of local neighbors, respectively.}
\resizebox{0.49\textwidth}{!}{
\begin{tabular}{ll | c| c | c | cc |  cc }
\toprule[0.5mm]

\multicolumn{2}{l|}{ }& &Time &Space & \multicolumn{2}{c|}{TSP100 }  & \multicolumn{2}{c}{TSP1000}\\ 
\multicolumn{2}{l|}{Method}  & \#Params &complexity&complexity&Avg.memory  & Time&  Avg.memory & Time\\

\midrule
\multicolumn{2}{l|}{POMO}   & \cellcolor[HTML]{D0CECE}\textbf{1.27M}& $\mathcal{O}(N^3)$& $\mathcal{O}(N^2)$& 1.62MB& 8s & 108.97MB &1.1m \\

\multicolumn{2}{l|}{ELG} & 1.27M& $\mathcal{O}(N^3+N^2k)$& $\mathcal{O}(N^2+Nk)$&2.63MB & 22s &126.57MB & 2m\\

\multicolumn{2}{l|}{ICAM} &2.24M& $\mathcal{O}(N^3)$& $\mathcal{O}(N^2)$& \cellcolor[HTML]{D0CECE}\textbf{0.89MB} &\cellcolor[HTML]{D0CECE}\textbf{5s} & \cellcolor[HTML]{D0CECE}\textbf{51.69MB}& \cellcolor[HTML]{D0CECE}\textbf{28s}\\

\bottomrule[0.5mm]
\end{tabular}
}
\label{table:gpu_memory_time}
\vskip -0.2in
\end{table}

\section{Ablation Study}
\label{sec:ablation_study}
In this section, we conduct a detailed ablation study and analysis to demonstrate the effectiveness and robustness of ICAM, mainly including: 
\begin{enumerate}
    \item \textbf{Comparison under consistent training setting} (see Appendix \ref{ablation:vst_comparison}); 
    \item \textbf{Effects of different components of adaptation function} (see Appendix \ref{ablation:adaptation_components}); 
    \item \textbf{Effects of adaptation function applied to different modules} (see Appendix \ref{ablation:different_modules}); 
    \item \textbf{Effects of different stages} (see Appendix \ref{ablation:different_stages}); 
    \item \textbf{Effects of deeper encoder} (see Appendix \ref{ablation:deeper_encoder});
    \item \textbf{The performance of POMO-Adaptation} (see Appendix \ref{ablation:pomo_adaptation}); 
\end{enumerate}

Please note that, unless stated otherwise, the results presented in the ablation study reflect the best result from multiple trajectories. We do not employ instance augmentation in the ablation study, and the performance of TSP instances is used as the primary criterion for evaluation. 

\section{Conclusion}
\label{sec:conclusion}
In this work, we have proposed a novel ICAM to improve large-scale generalization for RL-based neural routing solvers. We design a simple yet efficient instance-conditioned adaptation function to significantly improve the generalization performance of existing models with a small time and memory overhead. Further, the instance-conditioned information is more effectively incorporated into the whole neural solution construction process via a powerful yet low-complexity AAFM and the new compatibility calculation. The experimental results on various TSP, CVRP, and ATSP instances show that ICAM achieves promising generalization abilities compared to other representative methods.

In the future, we aim to develop more efficient inference strategies for ICAM, such as employing it as a repair operator within existing local search frameworks or developing more efficient instance augmentation techniques. We also plan to extend its applicability to routing problems with complex constraints by integrating more instance-conditioned information in $f(N,d_{ij})$, and non-routing problems by extending the definition of the adaptation function beyond geometric distance to capture domain-specific constraints, all while maintaining good performance.


 

\bibliographystyle{IEEEtran}
\bibliography{references}

\vspace{-6ex}

\begin{IEEEbiography}
[{\includegraphics[width=1in,height=1.25in,clip,keepaspectratio]{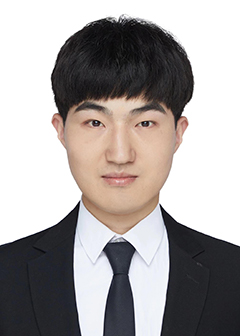}}]{Changliang Zhou} received his M.Eng. degree in software engineering from Qingdao University, Qingdao, China, in 2022. He is currently pursuing a Ph.D. at Southern University of Science and Technology in Shenzhen, China.

His main research interests include neural combinatorial optimization and its applications, with current focus on developing domain foundation models capable of solving a wide range of NP-hard COPs.
\end{IEEEbiography}

\vspace{-6ex}

\begin{IEEEbiography}
[{\includegraphics[width=1in,height=1.25in,clip,keepaspectratio]{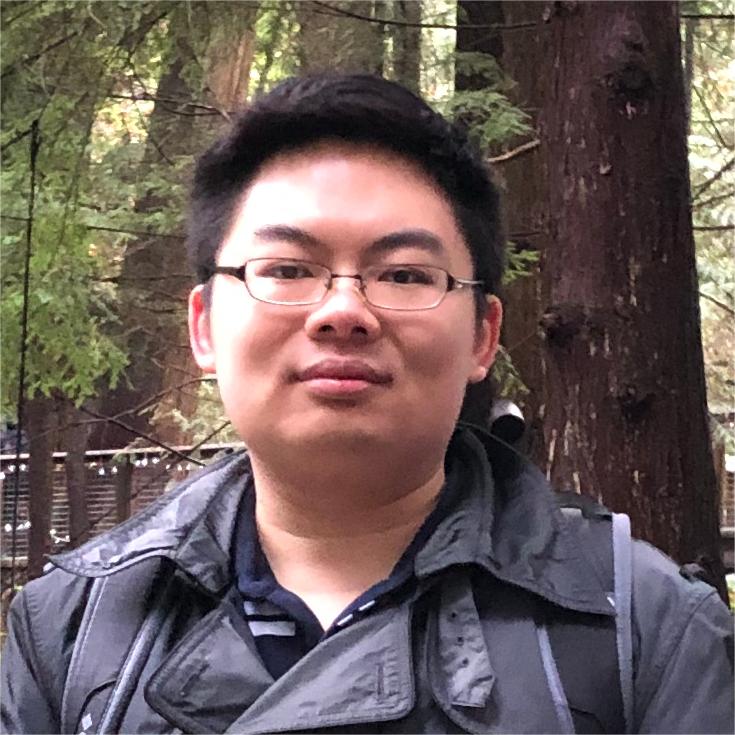}}]{Xi Lin} received his B.Sc. degree in Mathematics from South China University of Technology in 2013, his M.A. degree in Statistics from Columbia University in 2015, and his Ph.D. degree in Computer Science from City University of Hong Kong in 2020. He is currently a professor at the School of Mathematics and Statistics, Xi'an Jiaotong University, China. His main research interests include multi-objective optimization, learning-based optimization, and evolutionary computation. He has received several outstanding reviewer awards from ICML, ICLR, and TMLR. He currently serves as an Area Chair for NeurIPS and ICLR, and as an Action Editor for Transactions on Machine Learning Research.
\end{IEEEbiography}

\vspace{-4ex}

\begin{IEEEbiography}
[{\includegraphics[width=1in,height=1.25in,clip,keepaspectratio]{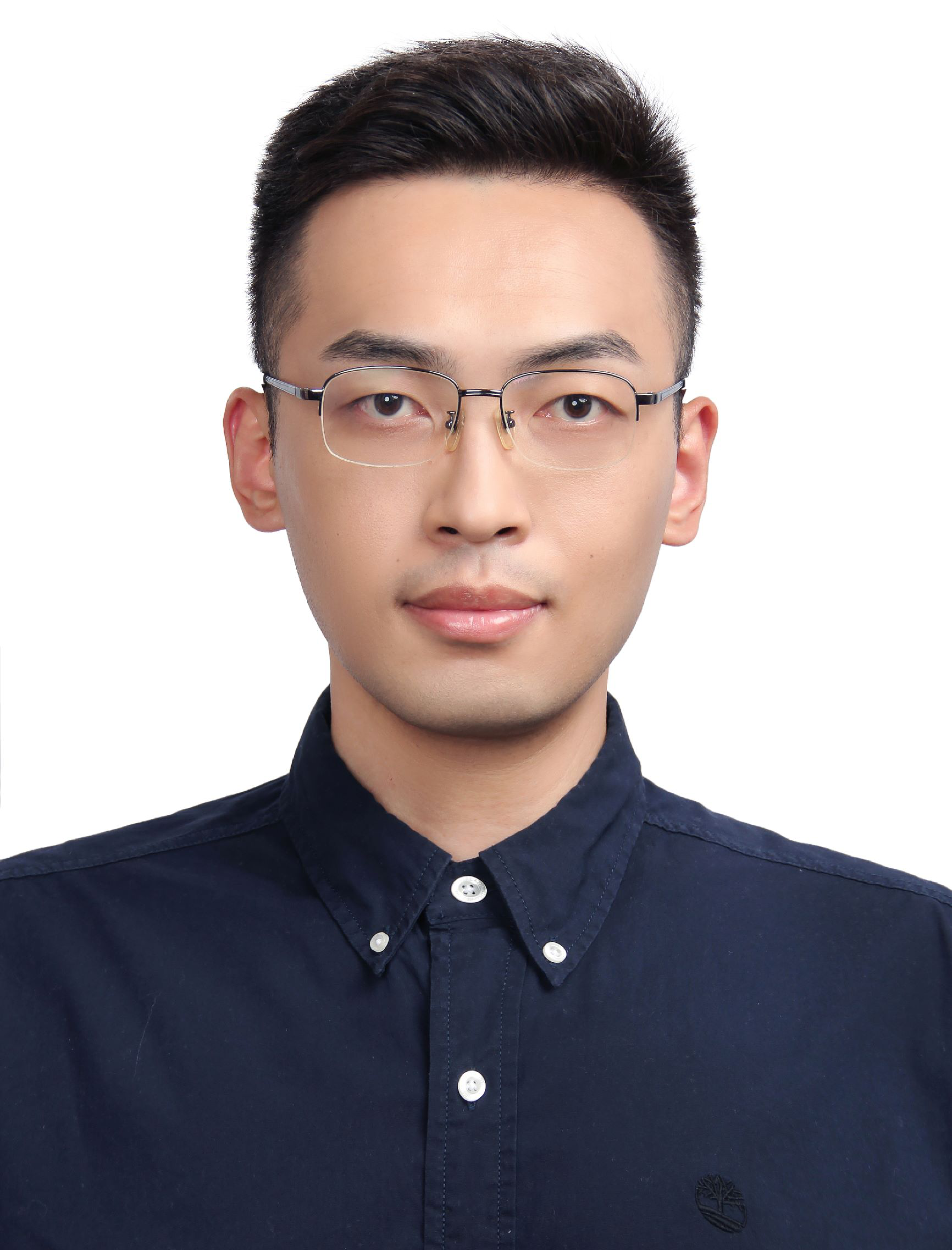}}]{Zhenkun Wang} (Senior Member, IEEE) received the Ph.D. degree in circuits and systems from Xidian University, Xi'an, China, in 2016. He is currently an Associate Professor with the School of Automation and Intelligent Manufacturing, Southern University of Science and Technology, Shenzhen, China. 

His research interests include AI-driven optimization, automated algorithm design, deep learning, and their applications.
\end{IEEEbiography}

\vspace{-4ex}

\begin{IEEEbiography}
[{\includegraphics[width=1in,height=1.25in,clip,keepaspectratio]{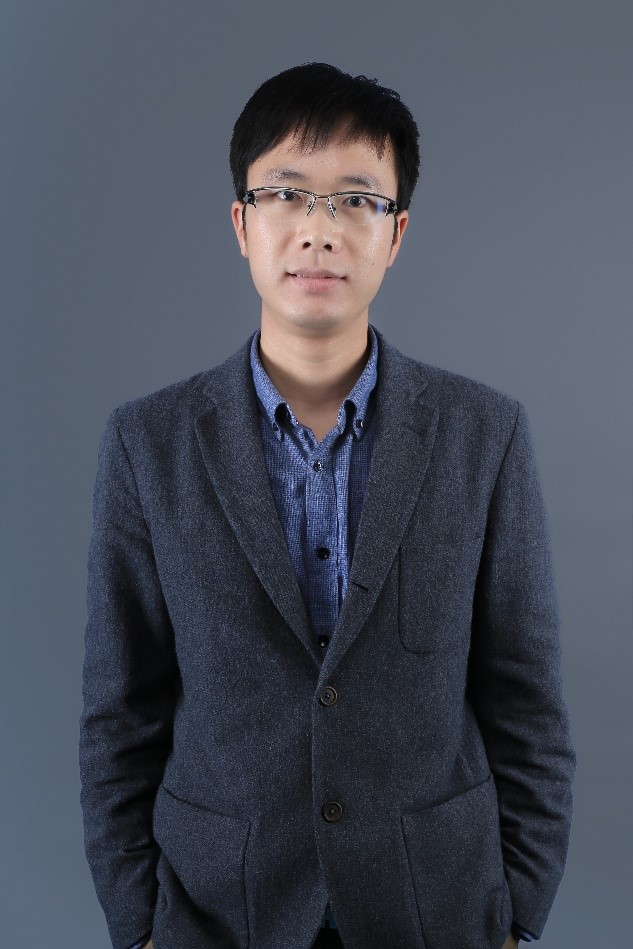}}]{Xialiang Tong} received his B. Sc. degree in telecommunication from Beijing Jiaotong University in 2009. He is currently a researcher at Huawei Noah’s Ark Lab, Shenzhen, China. His current research interests focus on efficient continual learning for large language models and their industrial applications. His previous research interests was centered on the applications of optimization algorithms in industries including logistics, manufacturing, ports, and energy.
\end{IEEEbiography}

\vspace{-4ex}

\begin{IEEEbiography}
[{\includegraphics[width=1in,height=1.25in,clip,keepaspectratio]{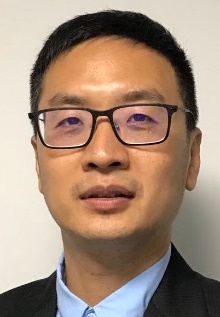}}]{Mingxuan Yuan} (Senior Member, IEEE) is currently a principal researcher of Huawei Noah’s Ark Lab. He also serves as the director of Hong Kong Noah’s Ark Lab. Before joining Huawei, he worked in HKUST as a post-doc researcher. He obtained his Ph.D degree from the Hong Kong University of Science and Technology. His research interests include Learning2Optimize and Applied AI Models. He has been the director of Huawei Applied AI Research Lab and has led several research projects including spatio-temporal data analysis, telecommunication data mining, enterprise intelligent, AI4EDA, AI-Solver and to Business Applied AI models.
\end{IEEEbiography}

\vspace{-4ex}

\begin{IEEEbiography}
[{\includegraphics[width=1in,height=1.25in,clip,keepaspectratio]{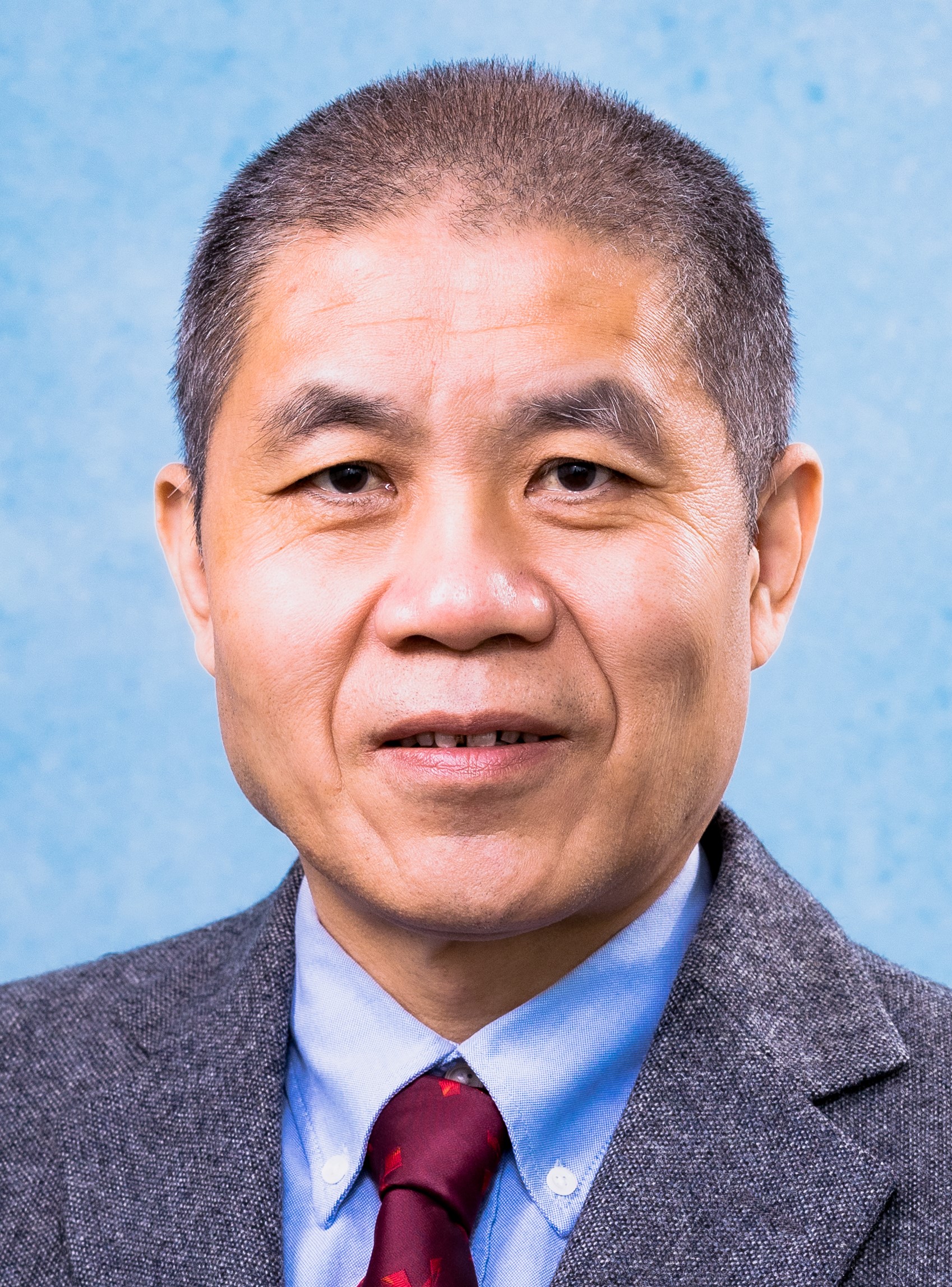}}]{Qingfu Zhang} (M’01-SM’06-F’17) received the BSc degree in mathematics from Shanxi University, China in 1984, the MSc degree in applied mathematics, and the PhD degree in information engineering from Xidian University, China, in 1991 and 1994, respectively.

He is a Chair Professor of Computational Intelligence at the Department of Computer Science, City University of Hong Kong. He heads a research group with a focus on metaheuristics and artificial intelligence. His MOEA/D algorithms have been widely studied and used in many application fields. He is a Web of Science highly cited researcher in Computer Science for eight times since 2016.

\end{IEEEbiography}


 



\clearpage
\appendices
This is the supplementary material for ``Instance-Conditioned Adaptation for Large-scale Generalization of Neural Routing Solver''. 

\section{Convergence curves of learnable biases}
\label{append:alpha_change}
Our proposed adaptation function is differentiable and forms part of both the AAFM and the subsequent compatibility module in the model. Its learnable parameters are simply trained end‑to‑end together with all other model parameters. In this work, we initialize the parameter to $1.0$ and use the gradient-based Adam optimizer to train all the model parameters (including the parameter of the adaptation function) for all experiments.

\begin{figure}[h!]
    \centering
    \begin{subfigure}{0.49\textwidth}
        \centering
        \includegraphics[width=\textwidth]{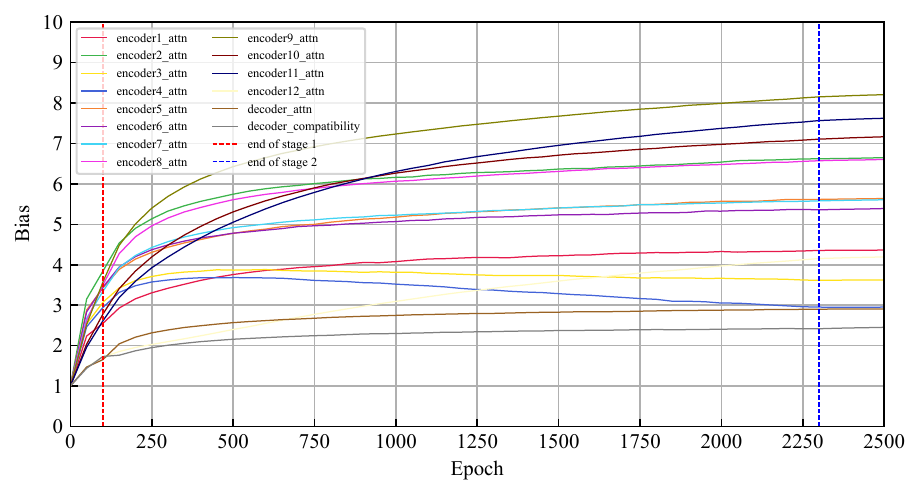} 
        \caption{TSP}
        \label{subfig:dist_hm_tsp100}
    \end{subfigure}
    \hfill
    \begin{subfigure}{0.49\textwidth}
        \centering
        \includegraphics[width=\textwidth]{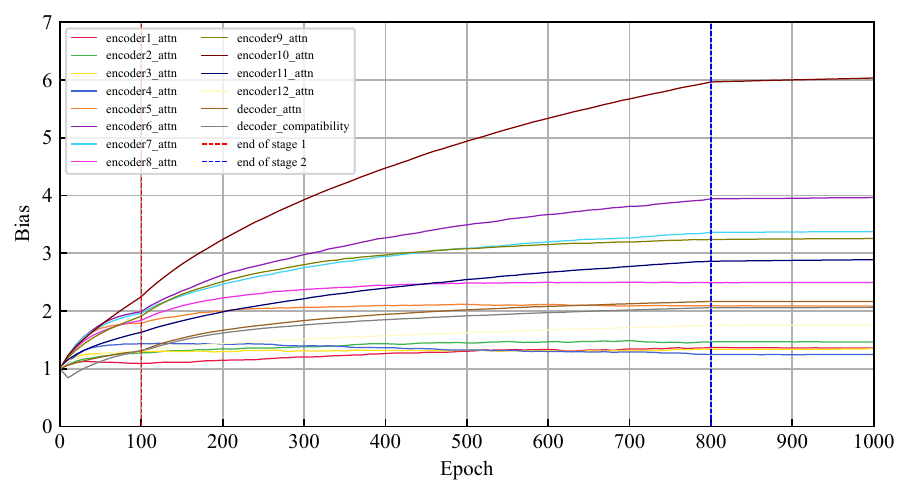}
        \caption{CVRP}
        \label{subfig:cosine_icam_tsp100}
    \end{subfigure}
    \caption{The convergence curve of learnable bias $\alpha$ throughout the training process for different problems (TSP and CVRP).}
    \label{fig:alpha_curve}
\end{figure}
The convergence curves of learnable bias $\alpha$ throughout the training process for TSP and CVRP are shown in Figure \ref{fig:alpha_curve}. According to the result, the parameter will converge to different values for different encoder and decoder layers.

\section{Detailed Ablation Study}
\label{append:ablation_study}

\subsection{Comparison under Consistent Training Settings}
\label{ablation:vst_comparison}
\begin{table}[h!]
\centering
\caption{Experimental results on TSPs and CVRPs under consistent training setting. Here, VST$n$ denotes this model is trained for $n$ epochs on varying-scale instances.}
\resizebox{0.48\textwidth}{!}{
\begin{tabular}{ll | ccc | ccc }
\toprule[0.5mm]

\multicolumn{2}{l|}{ }& \multicolumn{3}{c|}{TSP100 }  & \multicolumn{3}{c}{TSP1000}\\ 
\multicolumn{2}{l|}{Method}  & Obj. & Gap & Time  & Obj. & Gap & Time\\
\midrule
\multicolumn{2}{l|}{LKH3}   &7.7632 & 0.000\% & 56m & 23.1199 & 0.000\% & 8.2h \\
\midrule
\multicolumn{2}{l|}{POMO-Original} & \cellcolor[HTML]{D0CECE}\textbf{7.7915} &\cellcolor[HTML]{D0CECE}\textbf{ 0.365\%} & 8s & 32.8566 & 42.114\% & 1.1m\\
\multicolumn{2}{l|}{POMO-VST200}  & 7.9820& 2.818\% & 8s &25.8064 &11.620\% & 1.1m\\

\multicolumn{2}{l|}{ELG-Original}  &7.8128 & 0.638\% & 22s &25.7991 &11.588\% & 2m\\
\multicolumn{2}{l|}{ELG-VST200 } & 7.8429 & 1.027\% & 22s & 24.7273 & 6.953\% &  2m\\
\midrule
\multicolumn{2}{l|}{ICAM-VST20} & 7.8394  &0.982\% & 5s & 24.6161 &6.472\% &28s\\

\multicolumn{2}{l|}{ICAM-VST200} & 7.8284 &0.840\% & \cellcolor[HTML]{D0CECE}\textbf{5s} & \cellcolor[HTML]{D0CECE}\textbf{24.1331} &\cellcolor[HTML]{D0CECE}\textbf{ 4.382\% }&\cellcolor[HTML]{D0CECE}\textbf{28s}\\

\midrule
\midrule
 \multicolumn{2}{l|}{ }& \multicolumn{3}{c|}{CVRP100}  & \multicolumn{3}{c}{CVRP1000}\\
 \multicolumn{2}{l|}{Method}  & Obj. & Gap & Time  & Obj. & Gap & Time\\
\midrule
\multicolumn{2}{l|}{LKH3}   &15.6465 & 0.000\% & 12h & 37.0904 & 0.000\% & 7.1h \\
\midrule
\multicolumn{2}{l|}{POMO-Original} & \cellcolor[HTML]{D0CECE}\textbf{15.8368} & \cellcolor[HTML]{D0CECE}\textbf{1.217\%} & 10s & 143.1178 & 285.862\% & 1.2m\\
\multicolumn{2}{l|}{POMO-VST200}  &16.1019 & 2.911\% & 10s &40.1454 &8.237\% & 1.2m\\

\multicolumn{2}{l|}{ELG-Original}  &15.9855 & 2.166\% &34s  &42.0760 &13.442\% & 2.4m\\
\multicolumn{2}{l|}{ELG-VST200} &16.1121 & 2.975\% & 34s &39.7601 &  7.198\% &2.4m\\
\midrule
\multicolumn{2}{l|}{ICAM-VST20} & 16.0496 &2.576\% & 7s &  39.3221 &6.017\% &35s\\

\multicolumn{2}{l|}{ICAM-VST200} & 16.0240 &2.413\% & \cellcolor[HTML]{D0CECE}\textbf{7s} & \cellcolor[HTML]{D0CECE}\textbf{39.0220} & \cellcolor[HTML]{D0CECE}\textbf{5.208\%} &\cellcolor[HTML]{D0CECE}\textbf{35s}\\

\bottomrule[0.5mm]
\end{tabular}
}
\label{table:uniform_vst_pomo_elg_icam}
\vskip -0.2in
\end{table}
To improve the ability to be aware of scale, we implement a varying-scale training scheme. To ensure a fair comparison with baseline models, we conduct the consistent varying-scale training with 200 epochs (VST200) for both our proposed ICAM as well as the representative RL-based POMO and ELG models. The SL-based LEHD and BQ are not included in this experiment since it is difficult to obtain high-quality solutions for a large number of instances up to 500 nodes. As shown in \cref{table:uniform_vst_pomo_elg_icam}, our proposed varying-scale training (VST) method can also significantly improve the generalization performance of POMO and ELG. For example, ELG-VST200 can obtain a $6.9\%$ optimality gap on TSP1000 while the gap is $11.588\%$ for the original ELG. However, it should be emphasized that our proposed ICAM can achieve a better generalization after only $20$ epochs of varying-scale training. Given the substantial variations in patterns and geometric structures across different-scale routing instances, we argue that this stems from a better instance-conditioned adaptation of ICAM. These experimental results and analyses can be found in Section \ref{sec:capturing_features}.

\subsection{Effects of Different Components of Adaptation Function}
\label{ablation:adaptation_components}
In our adaptation function, except for the fundamental scale and pair-wise distance information, we additionally impose a learnable parameter. To better illustrate the effectiveness of this function, we conduct ablation experiments for the components, and the experimental results are shown in \cref{table:effect_components}. The results show that both a learnable parameter $\alpha$ and scale $N$ can significantly improve the model performance.
\begin{table}[h!]
\centering
\caption{Comparison between different component settings in the adaptation function.}
\resizebox{0.5\textwidth}{!}{
\begin{tabular}{ll | c | c | c | c }
\toprule[0.5mm]
\multicolumn{2}{l|}{ } & TSP100 & TSP200 & TSP500 & TSP1000  \\ 
\midrule
\multicolumn{2}{l|}{w/o learnable $\alpha$} & 0.546\% &  1.124\% & 2.785\% &  5.232\%\\
\multicolumn{2}{l|}{w/o scale} & 0.512\% &  0.866\% & 2.036\% &  4.236\%\\

\multicolumn{2}{l|}{w/ learnable $\alpha$ + scale} & \cellcolor[HTML]{D0CECE}\textbf{0.462\%} &\cellcolor[HTML]{D0CECE}\textbf{ 0.669\%} & \cellcolor[HTML]{D0CECE}\textbf{1.067\%} &  \cellcolor[HTML]{D0CECE}\textbf{1.907\%} \\

\bottomrule[0.5mm]
\end{tabular}
}
\label{table:effect_components}
\vskip -0.05in
\end{table}

\subsection{Effects of Adaptation Function Applied to Different Modules}
\label{ablation:different_modules}
\begin{table}[h!]
\centering
\caption{Comparison between effects of the adaptation function applied to different modules. Here, AFM denotes that AAFM removes the adaptation bias, and CAB is the compatibility with the adaptation bias.}
\label{table:detailed_incorporation_effect}
\resizebox{0.48\textwidth}{!}{
\begin{tabular}{ll | c | c | c | c }
\toprule[0.5mm]
\multicolumn{2}{l|}{ } & TSP100 & TSP200 & TSP500 & TSP1000  \\ 
\midrule
\multicolumn{2}{l|}{AFM} & 1.395\% &  2.280\% & 4.890\% &  8.872\%\\
\multicolumn{2}{l|}{AFM+CAB} & 0.956\% & 1.733\% & 4.081\% &  7.090\% \\
\multicolumn{2}{l|}{AAFM} & 0.514\% &  0.720\% & 1.135\% &  2.241\%\\
\multicolumn{2}{l|}{AAFM+CAB} & \cellcolor[HTML]{D0CECE}\textbf{0.462\%} &\cellcolor[HTML]{D0CECE}\textbf{ 0.669\%} & \cellcolor[HTML]{D0CECE}\textbf{1.067\%} &  \cellcolor[HTML]{D0CECE}\textbf{1.907\%} \\
\bottomrule[0.5mm]
\end{tabular}
}
\end{table}
Given that we apply the adaptation function to both the AAFM and the subsequent compatibility calculation, we conduct three different experiments to validate the efficacy of this function. The data presented in \cref{table:detailed_incorporation_effect} indicates a notable enhancement in the solving performance across various scales when instance-conditioned information is integrated into the model. This improvement emphasizes the importance of including detailed, fine-grained information in the model. It also highlights the critical role of explicit instance-conditioned information in improving the adaptability and generalization capabilities of RL-based models. In particular, the incorporation of richer instance-conditioned information allows the model to more effectively comprehend and address the challenges, especially in the context of large-scale problems.

\subsection{Effects of Different Stages}
\label{ablation:different_stages}
Our training is divided into three different stages, each contributing significantly to the overall effectiveness. The performance improvements achieved at each stage are detailed in \cref{table:effect_stages}. After the first stage, which uses only short training epochs, the model performs outstandingly with small-scale instances but underperforms when dealing with large-scale instances. After the second stage, there is a marked improvement in the ability to solve large-scale instances. By the end of the final stage, the overall performance is further improved. Notably, in our ICAM, the capability to tackle small-scale instances is not affected despite the instance scales varying during the training.
\begin{table}[h!]
\centering
\caption{Comparison of different training stages.}
\resizebox{0.48\textwidth}{!}{
\begin{tabular}{ll | c | c | c | c }
\toprule[0.5mm]
\multicolumn{2}{l|}{ } & TSP100 & TSP200 & TSP500 & TSP1000  \\ 
\midrule
\multicolumn{2}{l|}{After stage 1} & 0.514\% &  1.856\% & 7.732\% &  12.637\%\\
\multicolumn{2}{l|}{After stage 2} & 0.662\% &  0.993\% & 1.515\% &  2.716\%\\
\multicolumn{2}{l|}{After stage 3} & \cellcolor[HTML]{D0CECE}\textbf{0.462\%} &\cellcolor[HTML]{D0CECE}\textbf{ 0.669\%} & \cellcolor[HTML]{D0CECE}\textbf{1.067\%} &  \cellcolor[HTML]{D0CECE}\textbf{1.907\%} \\
\bottomrule[0.5mm]
\end{tabular}
}
\label{table:effect_stages}
\vskip -0.1in
\end{table}

\subsection{Effects of Deeper Encoder}
\label{ablation:deeper_encoder}
\begin{table}[h!]
\centering
\caption{Comparison of different encoder layers. Note that "L" represents encoder layers.}
\label{table:effects_encoder_layer}
\resizebox{0.48\textwidth}{!}{
\begin{tabular}{ll |ll | cc | cc }
\toprule[0.5mm]
\multicolumn{2}{l|}{ }&\multicolumn{2}{l|}{}& \multicolumn{2}{c|}{TSP100 } & \multicolumn{2}{c}{TSP1000}\\
\multicolumn{2}{l|}{Method}  &\multicolumn{2}{l|}{\#Params}  & Gap & Time  & Gap & Time\\
\midrule

\multicolumn{2}{l|}{ICAM-6L} &\multicolumn{2}{c|}{\cellcolor[HTML]{D0CECE}\textbf{1.15M}} & \cellcolor[HTML]{D0CECE}\textbf{0.442\%} & 5s  & 2.422\% & 28s\\

\multicolumn{2}{l|}{ICAM-12L} &\multicolumn{2}{c|}{2.24M} & 0.462\% & 5s  & \cellcolor[HTML]{D0CECE}\textbf{1.907\%} & 28s\\
\bottomrule[0.5mm]
\end{tabular}
}
\end{table}

We conduct an ablation study of ICAM with 6 and 12 layers, respectively. From these results in \cref{table:effects_encoder_layer}, we can see that a deeper encoder structure helps the model perform better in larger-scale instances. The ICAM-6L can already obtain promising results with fewer parameters. Furthermore, ICAM-12L can outperform ICAM-6L on large-scale instances. Due to our 12-layer encoder, ICAM-12L has more parameters than ICAM-6L. However, since the heavy encoder is only called once for the solution construction process, there is no obvious time difference between the models with 12-layer and 6-layer encoder.

\subsection{The Performance of POMO-Adaptation}
\label{ablation:pomo_adaptation}
We conduct an ablation study on the three-stage training for POMO equipped with our proposed adaption function. According to \cref{table:compare_pomo_twostage}, the adaption function and three-stage training scheme can significantly improve the generalization performance of POMO on large-scale problem instances. However, ICAM still performs better than POMO-Adaptation, both in terms of inference time and solution lengths.
\begin{table}[h!]
\centering
\caption{Comparison between different training stages of POMO with the adaptation function.}
\resizebox{0.49\textwidth}{!}{
\begin{tabular}{ll | ccc | ccc }
\toprule[0.5mm]
\multicolumn{2}{l|}{ }& \multicolumn{3}{c|}{TSP100 }  & \multicolumn{3}{c}{TSP1000}\\ 
\multicolumn{2}{l|}{Method}  & Obj. & Gap & Time  & Obj. & Gap & Time\\
\midrule
\multicolumn{2}{l|}{Concorde} &7.7632 & 0.000\% & 34m & 23.1199 & 0.000\% & 7.8h \\
\midrule
\multicolumn{2}{l|}{POMO-Original} & \cellcolor[HTML]{D0CECE}\textbf{7.7915} &\cellcolor[HTML]{D0CECE}\textbf{ 0.365\%} & 8s & 32.8566 & 42.114\% & 1.1m\\
\multicolumn{2}{l|}{POMO-Adaptation (Stage1)} & 7.9803 & 2.796\% & 9s & 26.9251  & 16.459\% &1.4m\\
\multicolumn{2}{l|}{POMO-Adaptation (Stage1,2)} & 8.0135  & 3.224\% & 9s & 24.6219 & 6.496\% &1.4m\\
\multicolumn{2}{l|}{POMO-Adaptation (Stage1,2,3)} & 7.9906 & 2.929\% & 9s &24.2849  & 5.039\% &1.4m\\
\midrule
\multicolumn{2}{l|}{ICAM (Stage1,2,3)} & 7.7991 & 0.462\% & \cellcolor[HTML]{D0CECE}\textbf{5s} &\cellcolor[HTML]{D0CECE}\textbf{ 23.5608} & \cellcolor[HTML]{D0CECE}\textbf{1.907\%} &  \cellcolor[HTML]{D0CECE}\textbf{28s}\\

\bottomrule[0.5mm]

\end{tabular}
}
\label{table:compare_pomo_twostage}
\end{table}

\section{Applicability to Real-world Systems}
\label{append:real_world}

\begin{figure*}[h!]
    \centering
    \begin{subfigure}{0.24\textwidth}
        \centering
        \includegraphics[width=\textwidth]{pictures/cities/nodes_map_Beijing_100.html.pdf} 
        \caption{Beijing-100}
    \end{subfigure}
    \hfill
    \begin{subfigure}{0.24\textwidth}
        \centering
        \includegraphics[width=\textwidth]{pictures/cities/nodes_map_Beijing_500.html.pdf} 
        \caption{Beijing-500}
    \end{subfigure}
    \hfill
     \begin{subfigure}{0.24\textwidth}
        \centering
        \includegraphics[width=\textwidth]{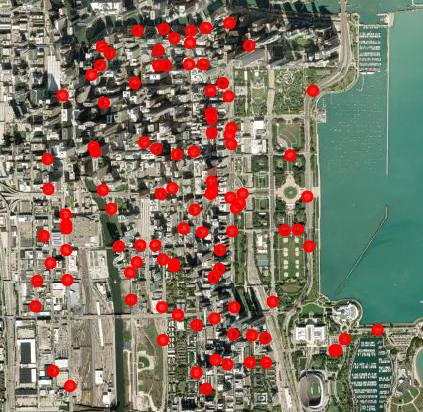} 
        \caption{Chicago-100}
    \end{subfigure}
    \hfill
    \begin{subfigure}{0.24\textwidth}
        \centering
        \includegraphics[width=\textwidth]{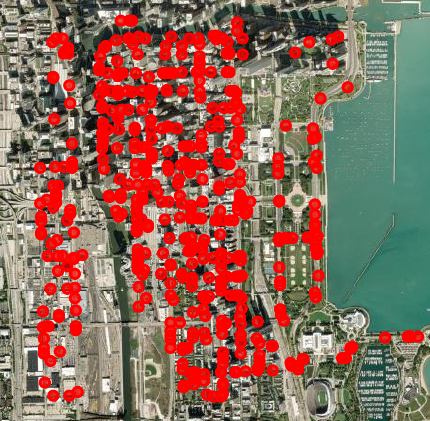} 
        \caption{Chicago-500}
    \end{subfigure}
    \hfill
    \begin{subfigure}{0.24\textwidth}
        \centering
        \includegraphics[width=\textwidth]{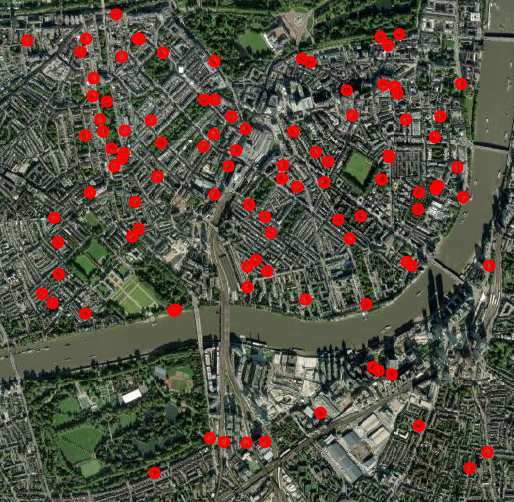} 
        \caption{London-100}
    \end{subfigure}
    \hfill
    \begin{subfigure}{0.24\textwidth}
        \centering
        \includegraphics[width=\textwidth]{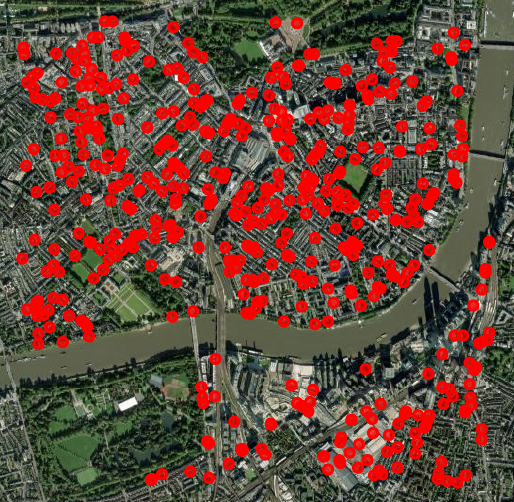} 
        \caption{London-500}
    \end{subfigure}
    \hfill
    \begin{subfigure}{0.24\textwidth}
        \centering
        \includegraphics[width=\textwidth]{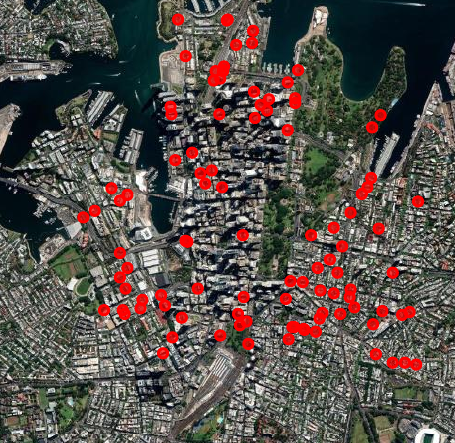} 
        \caption{Sydney-100}
    \end{subfigure}
    \hfill
    \begin{subfigure}{0.24\textwidth}
        \centering
        \includegraphics[width=\textwidth]{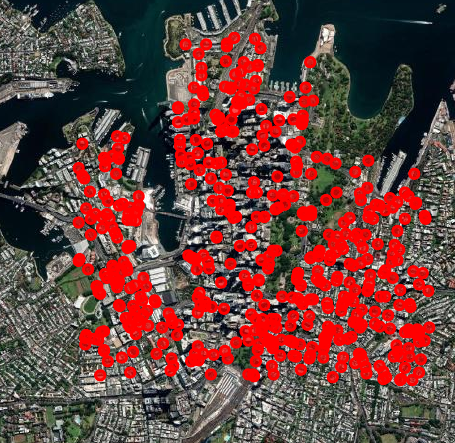} 
        \caption{Sydney-500}
    \end{subfigure}
    \hfill
    \caption{Visualization of instances sampled from different real-world cities. The notation 'City-$n$' indicates an instance with $n$ customer nodes sampled from the original map.}
    \label{fig:all_real_world_visual}
\end{figure*}

To further validate the effectiveness of ICAM in real-world systems, we have conducted a comprehensive case study using real-world road networks obtained from the OpenStreetMap (OSM) platform. We provide a visualization of instances sampled from different real-world cities in \cref{fig:all_real_world_visual}. For more analysis and experimental results, please refer to Section \ref{sec:real_world}.

\section{Capturing Instance-specific Features} 
\label{append:capturing_features}
\begin{figure*}[h!]
    \centering
    \begin{subfigure}{0.24\textwidth}
        \centering
        \includegraphics[width=\textwidth]{pictures/hm/tsp100_dist_matrix.pdf} 
        \caption{\centering Pair-wise distance \newline (TSP100)}
        \label{subfig:dist_hm_tsp100}
    \end{subfigure}
    \hfill
    \begin{subfigure}{0.24\textwidth}
        \centering
        \includegraphics[width=\textwidth]{pictures/hm/tsp100_icam_cosine.pdf}
        \caption{\centering Pair-wise similarity\newline (ICAM, TSP100)}
        \label{subfig:cosine_icam_tsp100}
    \end{subfigure}
    \hfill
     \begin{subfigure}{0.24\textwidth}
        \centering
        \includegraphics[width=\textwidth]{pictures/hm/tsp100_elg_cosine.pdf}
        \caption{\centering Pair-wise similarity\newline (ELG, TSP100)}
        \label{subfig:cosine_elg_tsp100}
    \end{subfigure}
    \hfill
     \begin{subfigure}{0.24\textwidth}
        \centering
        \includegraphics[width=\textwidth]{pictures/hm/tsp100_pomo_cosine.pdf}
        \caption{\centering Pair-wise similarity\newline (POMO,TSP100)}
        \label{subfig:cosine_pomo_tsp100}
    \end{subfigure}
    \hfill
    \begin{subfigure}{0.24\textwidth}
        \centering
        \includegraphics[width=\textwidth]{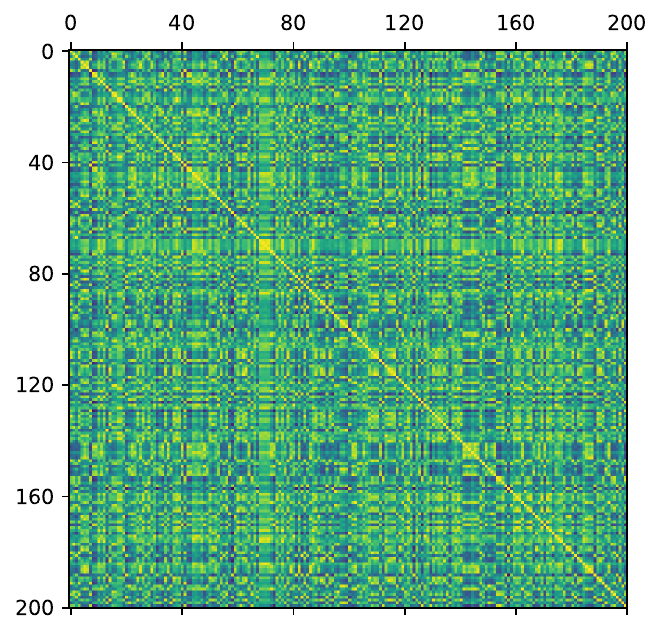} 
        \caption{\centering Pair-wise distance \newline (TSP200)}
        \label{subfig:dist_hm_tsp200}
    \end{subfigure}
    \hfill
    \begin{subfigure}{0.24\textwidth}
        \centering
        \includegraphics[width=\textwidth]{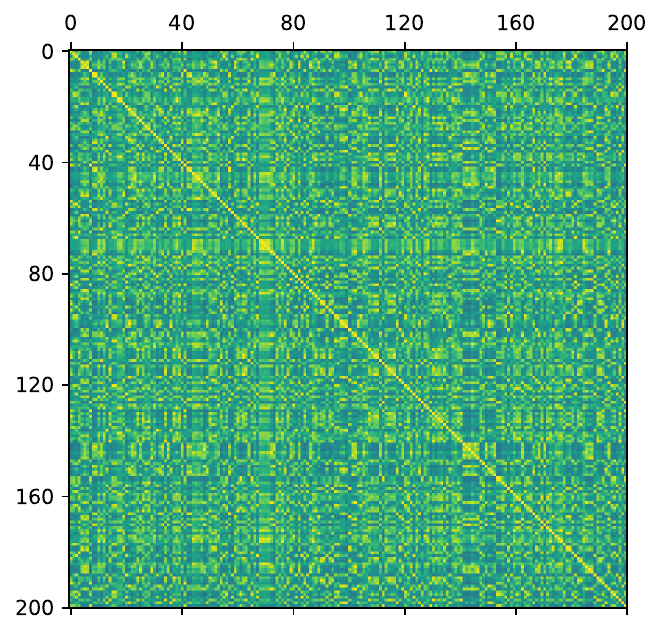}
        \caption{\centering Pair-wise similarity\newline (ICAM, TSP200)}
        \label{subfig:cosine_icam_tsp200}
    \end{subfigure}
    \hfill
     \begin{subfigure}{0.24\textwidth}
        \centering
        \includegraphics[width=\textwidth]{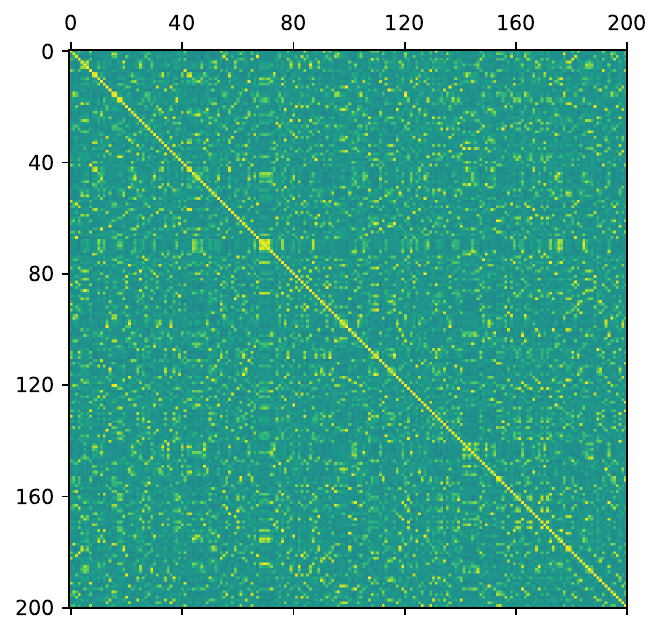}
        \caption{\centering Pair-wise similarity\newline (ELG, TSP200)}
        \label{subfig:cosine_elg_tsp200}
    \end{subfigure}
    \hfill
     \begin{subfigure}{0.24\textwidth}
        \centering
        \includegraphics[width=\textwidth]{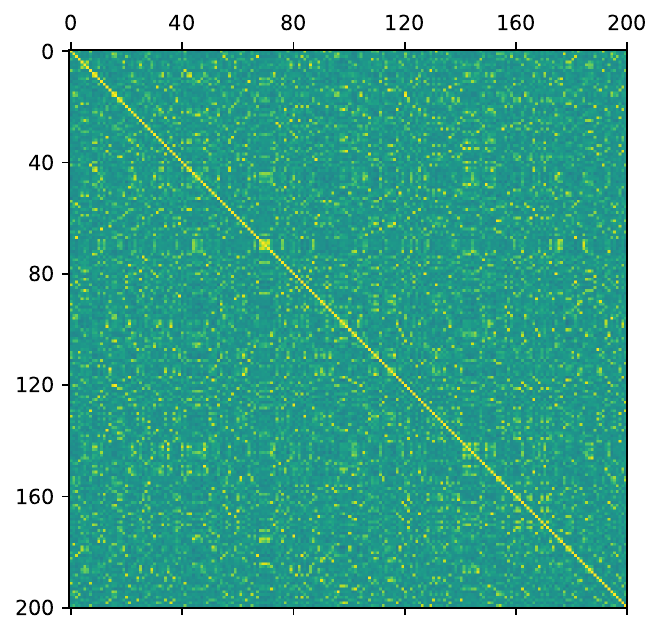}
        \caption{\centering Pair-wise similarity\newline (POMO,TSP200)}
        \label{subfig:cosine_pomo_tsp200}
    \end{subfigure}
    \hfill
    \begin{subfigure}{0.24\textwidth}
        \centering
        \includegraphics[width=\textwidth]{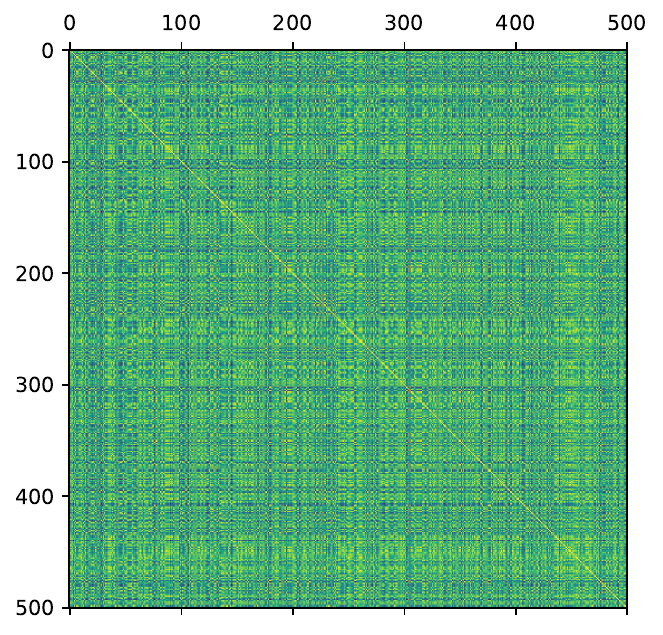} 
        \caption{\centering Pair-wise distance \newline (TSP500)}
        \label{subfig:dist_hm_tsp500}
    \end{subfigure}
    \hfill
    \begin{subfigure}{0.24\textwidth}
        \centering
        \includegraphics[width=\textwidth]{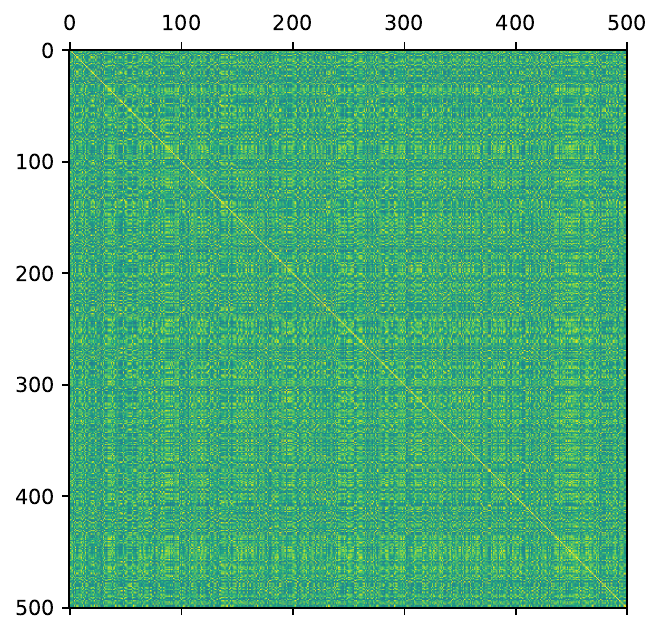}
        \caption{\centering Pair-wise similarity\newline (ICAM, TSP500)}
        \label{subfig:cosine_icam_tsp500}
    \end{subfigure}
    \hfill
     \begin{subfigure}{0.24\textwidth}
        \centering
        \includegraphics[width=\textwidth]{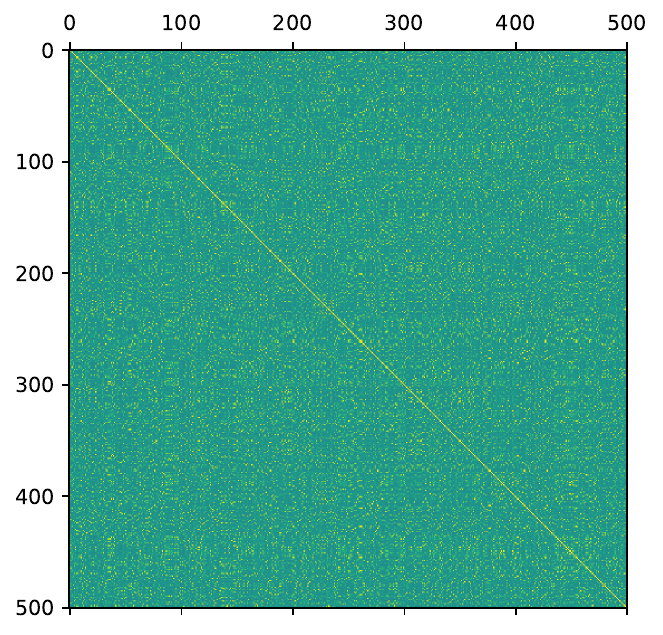}
        \caption{\centering Pair-wise similarity\newline (ELG, TSP500)}
        \label{subfig:cosine_elg_tsp500}
    \end{subfigure}
    \hfill
     \begin{subfigure}{0.24\textwidth}
        \centering
        \includegraphics[width=\textwidth]{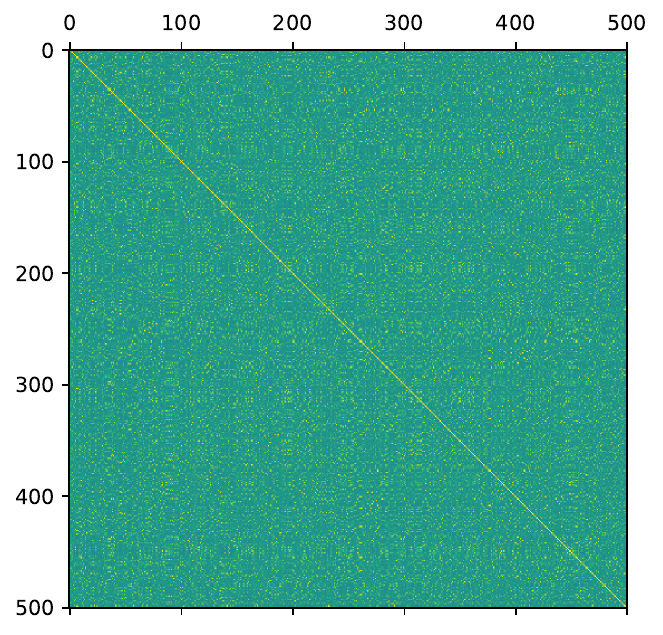}
        \caption{\centering Pair-wise similarity\newline (POMO,TSP500)}
        \label{subfig:cosine_pomo_tsp500}
    \end{subfigure}
    \hfill
    \begin{subfigure}{0.24\textwidth}
        \centering
        \includegraphics[width=\textwidth]{pictures/hm/tsp1000_dist_matrix.pdf} 
        \caption{\centering Pair-wise distance \newline (TSP1000)}
        \label{subfig:dist_hm_tsp1000}
    \end{subfigure}
    \hfill
    \begin{subfigure}{0.24\textwidth}
        \centering
        \includegraphics[width=\textwidth]{pictures/hm/tsp1000_icam_cosine.pdf}
        \caption{\centering Pair-wise similarity\newline (ICAM, TSP1000)}
        \label{subfig:cosine_icam_tsp1000}
    \end{subfigure}
    \hfill
     \begin{subfigure}{0.24\textwidth}
        \centering
        \includegraphics[width=\textwidth]{pictures/hm/tsp1000_elg_cosine.pdf}
        \caption{\centering Pair-wise similarity\newline (ELG, TSP1000)}
        \label{subfig:cosine_elg_tsp1000}
    \end{subfigure}
    \hfill
     \begin{subfigure}{0.24\textwidth}
        \centering
        \includegraphics[width=\textwidth]{pictures/hm/tsp1000_pomo_cosine.pdf}
        \caption{\centering Pair-wise similarity\newline (POMO,TSP1000)}
        \label{subfig:cosine_pomo_tsp1000}
    \end{subfigure}

    \caption{Comparison of cosine similarity between node embeddings generated by the encoders of different models and actual pair-wise distance with different scales. It is noteworthy that darker shades indicate lower similarity. If the node embeddings can successfully capture the instance-specific features, its similarity matrix should share some similar patterns with the normalized inverse distance matrix.}
    \label{fig:tsp_hm}
    \vskip -0.2in
\end{figure*}

We conduct a case study on TSP to demonstrate the instance-conditioned adaptation ability for different models, where the results are shown in \cref{fig:tsp_hm}. According to the results, the representative RL-based models (i.e., ELG and POMO) all fail to effectively capture instance-specific features in their node embeddings. On the other hand, our proposed ICAM can generate instance-conditioned node embeddings, of which the embedding correlation matrix shares similar patterns with the original distance matrix. These results clearly show that ICAM can successfully capture instance-specific features in its embeddings, which leads to its promising generalization performance. For more details, please see Section \ref{sec:capturing_features}.

\vfill

\end{document}